\pdfoutput=1
\documentclass[11pt]{article}

\usepackage{booktabs}
\usepackage{graphicx}
\usepackage{amsmath}
\usepackage[colorlinks=true,linkcolor=blue!55!black,citecolor=green!45!black,urlcolor=blue!55!black,pdfborder={0 0 0}]{hyperref}
\usepackage{xcolor}
\hypersetup{linkcolor=blue!55!black,citecolor=green!45!black,urlcolor=blue!55!black}
\usepackage[super,comma,sort&compress]{natbib}
\usepackage{geometry}
\usepackage{microtype}
\usepackage{amssymb}
\usepackage{multirow}
\usepackage{array}
\usepackage{authblk}
\usepackage{float}
\usepackage{placeins}   %
\usepackage{needspace} %
\usepackage{caption}    %
\usepackage{longtable}  %

\newcommand{\mainheading}[1]{%
  \subsection*{#1}\addcontentsline{toc}{subsection}{#1}}
\newcommand{\headingref}[1]{\hyperref[#1]{\nameref*{#1}}}
\newcounter{suppmethod}
\renewcommand{\thesuppmethod}{1.\arabic{suppmethod}}
\newcounter{suppdetail}[suppmethod]
\renewcommand{\thesuppdetail}{\thesuppmethod.\arabic{suppdetail}}
\newcommand{\suppmethodheading}[2]{%
  \refstepcounter{suppmethod}%
  \subsection*{\thesuppmethod\quad #1}\label{#2}%
  \addcontentsline{toc}{subsection}{\thesuppmethod\quad #1}}
\newcommand{\suppdetailheading}[2]{%
  \refstepcounter{suppdetail}%
  \subsubsection*{\thesuppdetail\quad #1}\label{#2}%
  \addcontentsline{toc}{subsubsection}{\thesuppdetail\quad #1}}
\newcommand{\suppmethodref}[1]{%
  \hyperref[#1]{Supplementary Method~\ref*{#1}}}
\hypersetup{bookmarksnumbered=true,bookmarksdepth=3}

\definecolor{bestcolor}{HTML}{CC0000}
\definecolor{secondcolor}{HTML}{0055AA}

\title{Health-System-Wide Mapping of Demographic Information in Routine Clinical Histopathology}

\author[1,2,$\dagger$]{Shengjia Chen}
\author[1,2]{Ruchika Verma}
\author[4]{Kevin Clare}
\author[1,2]{Paula Restrepo}
\author[1,2]{Jannes Jegminat}
\author[1,2]{Eugenia Alleva}
\author[3]{Kuan-lin Huang}
\author[4]{Aryeh Stock}
\author[4]{Brandon Veremis}
\author[1,2]{Thomas Fuchs}
\author[1,2,$\dagger$]{Gabriele Campanella}

\affil[1]{Windreich Department of Artificial Intelligence and Human Health,
  Icahn School of Medicine at Mount Sinai, New York, USA}
\affil[2]{Hasso Plattner Institute for Digital Health at Mount Sinai,
  Icahn School of Medicine at Mount Sinai, New York, USA}
\affil[3]{Mount Sinai Center for Transformative Disease Modeling,
  Icahn School of Medicine at Mount Sinai, New York, USA}
\affil[4]{Department of Pathology, Molecular and Cell-Based Medicine,
  Mount Sinai Health System, New York, USA}
\affil[$\dagger$]{Corresponding author: \texttt{shengjia.chen@icahn.mssm.edu} \\
                                      \texttt{gabriele.campanella@mssm.edu}}

\begin{document}
\maketitle

\begin{abstract}
The extent to which demographic-associated information is detectable across organs and disease states in routine clinical histopathology remains unclear. Here, we analyzed 1.3 million hematoxylin and eosin whole-slide images from 257,322 patients across 25 organs, using pathology foundation-model representations and multiple-instance learning to systematically map information associated with sex, age group, and self-reported race. Across 69 organ--demographic combinations, demographic attributes were broadly recoverable, with pan-organ mean AUROCs of 0.828 for sex, 0.733 for age group, and 0.662 for self-reported race, and substantial variation across organs. Demographic-class balancing produced relatively small changes in performance, while evaluation within diagnosis-text clusters showed modest attenuation. Under training with diagnosis-based grouping, demographic-associated signals remained detectable across multiple organs, including sex in gastric tissue (AUROC 0.865), age group in breast (0.816), and self-reported race in skin (0.728). Literature-guided analysis of these settings generated candidate histomorphological and contextual explanations for further investigation. Together, these findings provide a systematic map of demographic-associated information in routine histopathology, informing investigations of its sources, assessments of potential shortcut learning, and the development of more interpretable computational pathology models.
\end{abstract}

\section{Introduction}

Despite more than a century of microscopic examination of human tissue, it
remains unknown whether demographic-associated morphologic variation is
systematically detectable across organs and disease states.
Scattered observations suggest that such variation exists. Sex-associated
nuclear features were described in neurons and neutrophils more than 70 years
ago\cite{barr1949nuclear,davidson1954drumstick}, and machine-learning models
have since estimated chronological age from the histology of healthy tissues
and brain\cite{gerstein2025imageqtl,braun2023histopathologic}. More broadly,
structural aging mapped from hematoxylin and eosin (H\&E) histology of 40
post-mortem tissues in the Genotype-Tissue Expression project followed tissue-specific, nonlinear
trajectories, with coordinated deterioration across
organs\cite{yadav2026pathstar}. Race-associated histopathologic and molecular
differences have also been reported in organs including skin and
prostate\cite{williams2023skin,bigler2011prostate,awasthi2021prostate}.
These observations establish that demographic-associated variation can have
histologic correlates, but the evidence remains fragmented across demographic
attributes, organs, and clinical settings. It is therefore unknown how broadly
sex, age, and self-reported race are associated with detectable information in
routine clinical histopathology across tissues and disease states.

Deep learning provides a means to investigate this question at a scale and
sensitivity not readily achievable through conventional morphologic
assessment. Computational pathology (CPath) models applied to digitized
H\&E whole-slide images (WSIs) can identify information
associated with disease and tumor grade\cite{campanella2019clinical,bulten2022panda},
molecular biomarkers and outcomes\cite{kather2019msi,campanella2025eagle,volinsky2024hector}, and cellular
phenotypes\cite{graham2019hovernet,horst2024cellvit}. Foundation models (FMs)
pretrained on large collections of digitized pathology images\cite{chen2024uni,xu2024gigapath,vorontsov2024virchow,ding2025titan,lu2023conch}
have further enabled representations that transfer across tasks and
centers\cite{campanella2025benchmark,neidlinger2025benchmark}. The ability of
these models to recover complex information from routine H\&E images raises a
fundamental question beyond disease prediction: what information about the
patient is encoded in tissue images, including information that may not be
apparent to human observers?

Evidence from medical imaging suggests that demographic information can indeed
be recovered from images and learned representations\cite{adleberg2022predicting}. In radiology, self-reported race can be
predicted across imaging modalities and external cohorts even when the
relevant image features are not apparent to human observers\cite{gichoya2022ai}; medical-imaging FMs likewise encode sex, age, and race
in their representations\cite{joo2025vlfoundation}. Evidence in
histopathology is more limited. Vaidya et al.\ showed that representations
learned for breast and lung cancer subtyping and glioma \textit{IDH1}
mutation prediction retained information about patient
race\cite{vaidya2024demographic}. In our prior work in dermatopathology,
self-reported race remained predictable under the evaluated
disease-distribution controls, with attention and ablation analyses
implicating the epidermis as one source of predictive
signal\cite{chen2025saif_skin}; pigmentation-related epidermal features are
one plausible candidate correlate\cite{williams2023skin}. Together, these
findings indicate that demographic information can be present in medical
images and learned representations, but they do not establish how broadly
such information is detectable across routine histopathology or which
histologic features carry it.

Establishing the extent of demographic predictability also has important
implications for the development and evaluation of CPath models. If
demographic attributes are recoverable from H\&E images or their learned
representations, this information is potentially available to models even
when demographic variables are not explicitly provided as inputs. Models may
then exploit demographic-associated information when it correlates with a
prediction target, creating the possibility of shortcut learning\cite{ongly2024shortcut,jones2024causal}. Performance disparities have been
documented in medical imaging and CPath\cite{joo2025vlfoundation,yang2024limits,vaidya2024demographic}; a pan-cancer
analysis, for example, reported significant disparities across self-reported
race, gender, or age groups in 29.3\% of diagnostic
tasks\cite{lin2025fairpath}. Self-reported race may serve as an imperfect proxy for genetic ancestry,
while also capturing social, environmental, and healthcare-related factors. Demographic predictability does not by itself
demonstrate that a clinical model uses demographic information, just as an
observed performance disparity does not establish demographic proxy use.
Rather, determining where demographic information is recoverable identifies
settings in which such use is possible and can be tested directly.

A central challenge is that demographic predictability need not arise from
intrinsic tissue morphology. Demographic groups can differ in disease
prevalence, specimen and case mix, healthcare context, and other cohort
characteristics, while tissue processing, staining, scanning, and image
acquisition can introduce additional correlations\cite{howard2021impact,ongly2024shortcut,jones2024causal}. The Cancer Genome Atlas
illustrates this problem: contributing sites differ in clinical and demographic
composition, while their digital slides carry site-specific visual signatures
arising from both biological differences and non-biological batch effects\cite{biased2023tcga,howard2021impact,vaidya2024demographic}. Deep learning
models can recover the submitting site from digitized slide images, and these
signatures remain detectable after commonly used stain normalization and
augmentation\cite{howard2021impact,huang2025flex}. When the same cohort
structure is represented in training and validation data, such associations
can remain hidden and produce optimistic estimates of performance that
deteriorate when site, case mix, or acquisition conditions change\cite{ongly2024shortcut,oakden2020hidden,jones2024causal,howard2021impact}.
Consequently, mapping demographic-associated image information requires
testing whether predictability persists when major sources of cohort
structure, particularly demographic-class imbalance and diagnosis context,
are constrained. Even when it persists, predictability alone does not
identify its causal source; residual signals may reflect tissue-associated
features, clinical or sampling context, acquisition effects, or combinations
of these factors.

Here, we address a fundamental question: to what extent are basic demographic
attributes encoded in tissue morphology across the human body? To that end, we
analyzed 1,302,440 H\&E WSIs from 257,322 patients across 25 organs
in the Mount Sinai Health System to systematically map demographic-associated
information in routine clinical histopathology. Using pathology FM features
and attention-based multiple-instance learning (MIL), we evaluated sex, age
group, and self-reported race across 69 organ--demographic combinations.
Complementary controls assessed demographic-class exposure balancing,
diagnosis-constrained versus non-stratified grouped training, and
within-diagnosis evaluation. These comparisons characterize the sensitivity
of demographic predictability to the evaluated controls without partitioning
the signal into causal components. Selected residual signals were further examined through targeted literature
searching to generate testable hypotheses for subsequent morphologic
investigation. This study maps where demographic information is image-predictable, where it persists under diagnosis and cohort controls, and which residual signals
warrant further morphologic investigation. It thereby provides a systematic
foundation for studying demographic-associated variation in human
histopathology and its implications for downstream CPath models.

\section{Methods}
\label{sec:methods}

\mainheading{Study design and cohort}
\label{sec:methods-cohort}

This retrospective cohort included all available digitized routine diagnostic
H\&E WSIs from the Mount Sinai Health System (MSHS) through October 2025,
together with linked pathology reports and clinical metadata. Cases without digitized H\&E WSIs were excluded. 
The finalized cohort comprised 257,322 patients, 327,483 pathology cases, 731,972 specimens, and
1,302,440 WSIs. The data followed a hierarchical structure: each patient could
contribute one or more cases, each case with  one or more specimens, and each
specimen with one or more WSIs. De-identified identifiers preserved these links.
Routine staining and scanning were centralized within MSHS. This study was approved by the Institutional Review Board at the
Icahn School of Medicine at Mount Sinai (Protocol STUDY-20-00338). Informed consent was waived under the approved protocol.

\mainheading{Report extraction and organ grouping}
\label{sec:methods-report-extraction}

Model development and evaluation were stratified by specimen organ. Pathology
reports were linked to cases and parsed at the specimen level using
OpenAI GPT-5 mini\cite{openai2025gpt5systemcard}, accessed through Azure OpenAI.
Schema-constrained outputs recorded each specimen's anatomic source,
normalized organ label, and verbatim diagnosis text. The
extraction protocol is summarized in Supplementary
Note~\ref{suppnote:diagnosis-extraction}.

Extracted organ labels were normalized into 25 groups selected for clinical
interpretability and sufficient data for model training and evaluation. Closely
related anatomic sites were consolidated where appropriate, including
colon/rectum, bone/joint, and soft tissue. Specimens not mapped to these groups
were assigned to an Other category and excluded from pan-organ analyses. The
final organ set included bladder, bone marrow, bone/joint, brain,
breast, cervix, colon/rectum, esophagus, gallbladder, gastric, kidney, liver, lung,
lymph node, mouth, ovary, pancreas, placenta, prostate, sinus, skin, small bowel, soft tissue, thyroid, and uterus.
For the cohort overview, specimen-level diagnosis text was visualized
using UMAP and classified into four broad categories (neoplastic,
normal, inflammatory, and other) using rule-based text mapping
(Fig.~\ref{fig:framework}B). Details are provided in
Supplementary Note~\ref{suppnote:dx-spectrum}.

\mainheading{Demographic labels}
\label{sec:methods-labels}
Eligibility was determined separately for each demographic target based on the availability of a valid label. Sex was recorded at the patient level and harmonized to female and male; indeterminate or unknown labels were excluded. Sex prediction was evaluated across 19 organ groups, excluding prostate, breast, cervix, ovary, uterus, and placenta. Self-reported race was recorded at the patient level and harmonized to five mutually exclusive categories: White, Black or African American, Hispanic or Latino, Asian, and Other. Other included Native Hawaiian or Other Pacific Islander, American Indian or Alaska Native, and Other Race; missing or unmappable values were excluded. Age at the pathology encounter was calculated as case receive year minus birth year and grouped into 20--34, 35--44, 45--54, 55--64, 65--74, and $\geq$75 years. Cases with missing age or age below 20 years were excluded from age-group analyses.
\mainheading{Diagnosis embeddings and clustering}
\label{sec:methods-text-embedding}

Diagnosis text embeddings were used to group specimens with similar diagnostic contexts for
stratified evaluation of demographic predictability and the construction of cluster-constrained training groups. Specimen-level diagnosis text was encoded using OpenAI
\texttt{text-embedding-3-large}, producing 3,072-dimensional embeddings. Embeddings were computed at the specimen level because a single pathology case may contain multiple
specimens from different anatomic sites with distinct diagnoses. For each organ, we applied
$k$-means clustering to the diagnosis embeddings. The number of clusters was selected
separately from an organ-specific candidate grid using silhouette scores. 
$K$-means was fit once per organ, over all eligible specimens for that organ. The resulting assignments were fixed across all
Monte Carlo cross-validation (MCCV) folds. 

\mainheading{Controlled training regimes}
\label{sec:methods-training-regimes}

The overall study design is summarized in Fig.~\ref{fig:results}A.
Observed-distribution training (S1) sampled cases according to the observed
demographic-class distribution. Demographic-balanced training (S2) used
case-level resampling to balance demographic-class exposure, using the S1
epoch-size target. Diagnosis-context-constrained grouped training (S3)
constructed demographic-balanced groups within diagnosis clusters in each
fold's training pool. Retained classes were oversampled to the largest class
before assembling groups containing one case from each retained class.

For group construction, specimen-level cluster assignments were mapped to
cases by majority vote, with ties resolved by the first tied cluster encountered
in specimen row order. The random grouped-training control (S3$_\text{RND}$)
applied the same demographic balancing procedure to the full training pool
without diagnosis-context constraints. This comparison assessed the additional
association of diagnosis-context constraints under grouped training; it did
not isolate data size or guarantee identical training exposure. Detailed
group-construction, class-retention, and epoch-exposure rules are provided in
\suppmethodref{supp:training-implementation}. All regimes shared the same
model architecture, prediction unit, and cross-validation design.

\mainheading{Model development}
\label{sec:methods-model-development}

For each case and organ target, eligible H\&E slides from organ-matched
specimens were combined into one case-level MIL bag, capped at 30 slides
and 100,000 tiles. For cases exceeding the 30-slide cap, slides were randomly subsampled
at each training epoch using a seeded random procedure.
At least one organ-matched specimen with linked H\&E WSIs
was required for inclusion.
GrandQC\cite{weng2024grandqc} excluded background, pen marks, and artifacts
before tile extraction at 20$\times$ magnification. The in-house small
vision transformer (ViT-small) pathology foundation model SP22M
\cite{campanella2023sp22m} generated 384-dimensional tile embeddings
without task-specific fine-tuning.

Gated multi-head attention MIL used
class-specific attention heads\cite{ilse2018attention,lu2021data}.
Models were trained with AdamW, a learning rate of $10^{-5}$,
10 epochs of linear warm-up, and cosine decay to $10^{-6}$ over
50 epochs. S1 and S2 processed one case at a time and computed case-level
losses, whereas S3 and its random grouped-training control summed case-level
losses across group members before gradient accumulation and optimizer updates. Detailed regime-specific loss aggregation,
epoch-exposure formulas, and group caps are provided in
\suppmethodref{supp:training-implementation}.

\mainheading{Evaluation and statistical analysis}
\label{sec:methods-evaluation}

Performance was estimated using 10-fold MCCV, with an 80/20
train-validation split per fold. Sex and self-reported race used
patient-level splits, whereas age group used case-level splits
to account for age variation across encounters. Evaluation was
performed at the case level. All models were evaluated at the
final training epoch (epoch 50).

The primary metric was macro one-vs-rest area under the receiver
operating characteristic curve (AUROC), reported as mean $\pm$
standard deviation (SD) across folds. Regime comparisons emphasized
paired AUROC differences and their consistency across organs and folds.
Paired $t$-tests were performed across matched MCCV folds, with
Benjamini--Hochberg (BH) adjustment applied separately within each
demographic target and regime comparison. Statistical results are provided
in Supplementary Table~\ref{tab:supp-regime-statistics}.

As an additional evaluation analysis, the existing S1 and S2 predictions
were evaluated within diagnosis-text clusters using the same matched
validation cases as pooled evaluation. Within-cluster AUROCs were aggregated
using case-count weighting and compared with pooled AUROC. This analysis
introduced neither a new training regime nor a prediction target and was
distinct from diagnosis-context-constrained S3 training. Clusters containing fewer than
two target classes were excluded. The same evaluation procedure
was applied across all organ and demographic combinations,
including the breast case study. Detailed definitions and
eligibility criteria are provided in
\suppmethodref{supp:cluster-evaluation}.

\mainheading{Literature searching and hypothesis generation}
\label{sec:methods-hypothesis-map}

Targeted large language model (LLM)-assisted literature searching was conducted
for the three selected organ--demographic combinations to identify potential
histomorphologic and clinical explanations for residual demographic
predictability. Using OpenAI \texttt{gpt-6-astra} (accessed 24 September 2026),
a standardized prompt was used to identify relevant studies, including
supporting, contradictory, and null findings. Candidate explanations were
checked in a separate model-assisted pass against retrieved abstracts or
accessible full texts for citation accuracy, evidence support, and
applicability to the study population and tissue context. The resulting
literature-supported hypotheses were summarized to guide subsequent
morphologic investigation. The search protocol and source verification
procedures are provided in \suppmethodref{supp:hypothesis-map}.

\section{Results}
\label{sec:results}

\mainheading{Cohort characteristics}
\label{sec:results-data}

\begin{figure}[htbp]
  \centering
  \includegraphics[width=\textwidth,height=0.75\textheight,keepaspectratio]{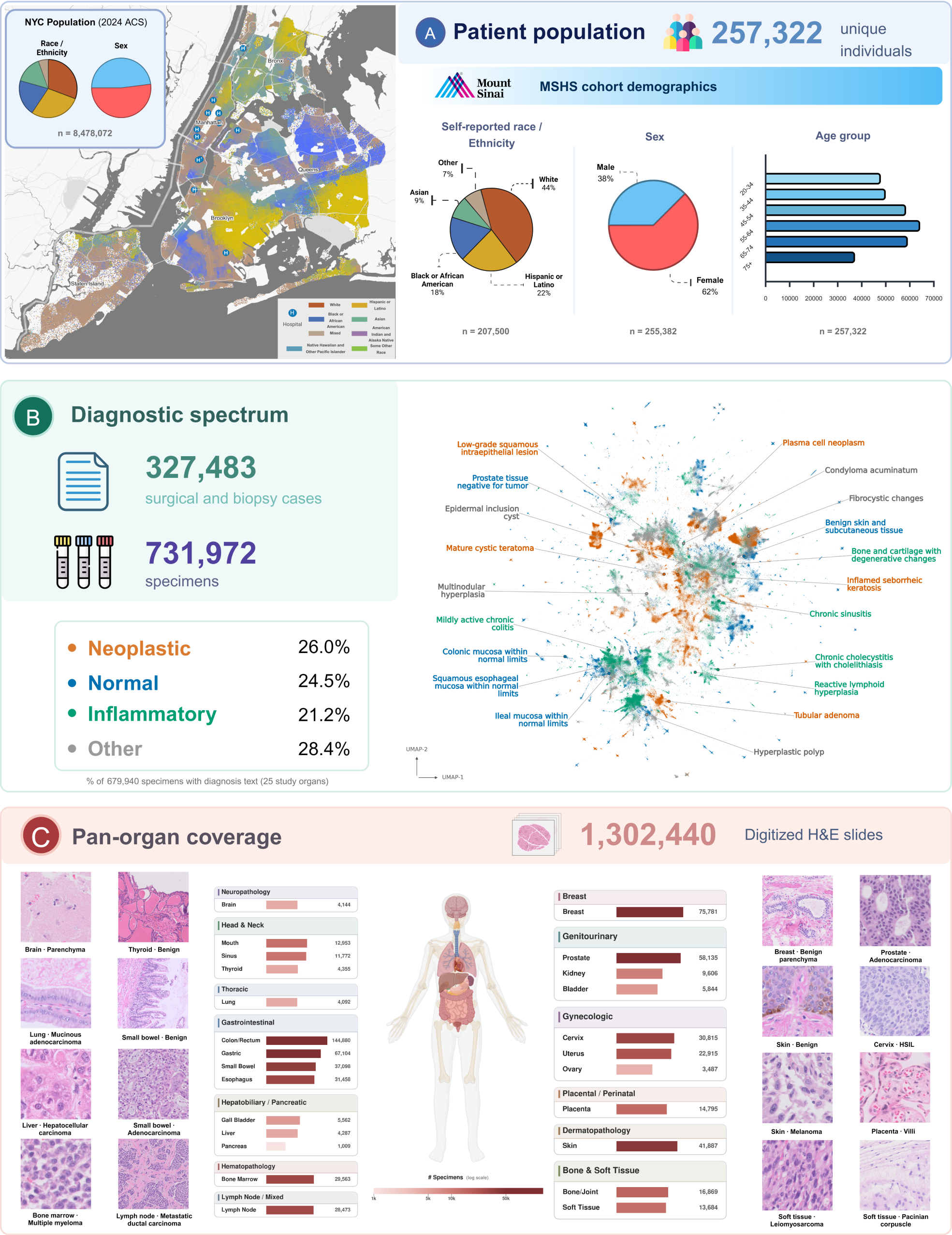}
  \caption{%
    \textbf{Scale and diversity of the MSHS routine histopathology cohort.}
    \textbf{(A) Population context.}~NYC geography and cohort demographics.
    ACS percentages use all residents: Hispanic/Latino of any race and
    non-Hispanic White, Black, and Asian; remaining categories differ from
    cohort Other. Cohort age is case-level.
    \textbf{(B) Diagnostic spectrum.}~Diagnosis-text UMAP and percentages for
    679,940 specimens from 25 organs. Colours: neoplastic, orange; normal,
    blue; inflammatory, green; other (including indeterminate), grey.
    Labels identify diagnoses; dense islands reflect templated sign-out text,
    not prevalence.
    \textbf{(C) Pan-organ coverage.}~The cohort comprises 1,302,440 digitized
    H\&E slides. Bars show specimen counts across 25 organs grouped into 13
    pathology subspecialty divisions, on a shared logarithmic scale (color
    saturation at $\geq$50,000 specimens). Sixteen illustrative H\&E images
    highlight histological diversity across organs.
    Created in BioRender. Chen, S. (2026) \url{https://BioRender.com/pc3001s}
  }
  \label{fig:framework}
\end{figure}

The study cohort comprised 257,322 patients, 327,483 surgical and biopsy cases,
731,972 pathology specimens, and 1,302,440 digitized H\&E WSIs from MSHS
in New York City (NYC) through October 2025 (Fig.~\ref{fig:framework}). Among patients with known sex ($n=255{,}382$), 159,411 (62.4\%) were female
and 95,971 (37.6\%) male. Among patients with known self-reported
race/ethnicity ($n=207{,}500$), 90,761 (43.7\%) were White, 47,220 (22.8\%)
Hispanic or Latino, 36,388 (17.5\%) Black or African American, 18,457 (8.9\%)
Asian, and 14,674 (7.1\%) Other. In the 2024 NYC ACS population
($n=8{,}478{,}072$), provided as an external demographic reference and not
included in the study cohort, 30.7\% were non-Hispanic White, 28.7\% Hispanic or
Latino, 20.0\% non-Hispanic Black or African American, and 15.1\%
non-Hispanic Asian\cite{acs2024_nyc}. Among cases with known age
($n=318{,}174$), 48,126 (15.1\%) were aged 20--34 years, 50,160 (15.8\%)
35--44, 58,635 (18.4\%) 45--54, 64,503 (20.3\%) 55--64, 59,356 (18.7\%)
65--74, and 37,394 (11.8\%) $\geq$75 (Fig.~\ref{fig:framework}A).
Full distributions and missing-label categories are reported in Supplementary
Table~\ref{tab:cohort}.

The diagnostic spectrum included neoplastic, normal, inflammatory, and
other findings across the study organs. Among the 679,940 mapped specimens,
26.0\% were neoplastic,
24.5\% normal, 21.2\% inflammatory, and 28.4\% other, including indeterminate
findings (Fig.~\ref{fig:framework}B; Supplementary
Note~\ref{suppnote:dx-spectrum}). Specimens spanned 25 organ groups organized into 13 pathology subspecialty
divisions. The largest groups were colon/rectum (144,880 specimens),
breast (75,781), gastric
(67,104), and prostate (58,135)
(Fig.~\ref{fig:framework}C; Supplementary Table~\ref{tab:supp-cohort-scale}).
Per-organ demographic composition is reported in Supplementary
Table~\ref{tab:supp-demographic-composition}.

\mainheading{Pan-organ demographic predictability}
\label{sec:results-s1}

\begin{figure}[htbp]
  \centering
  \includegraphics[width=\textwidth,height=0.74\textheight,keepaspectratio]{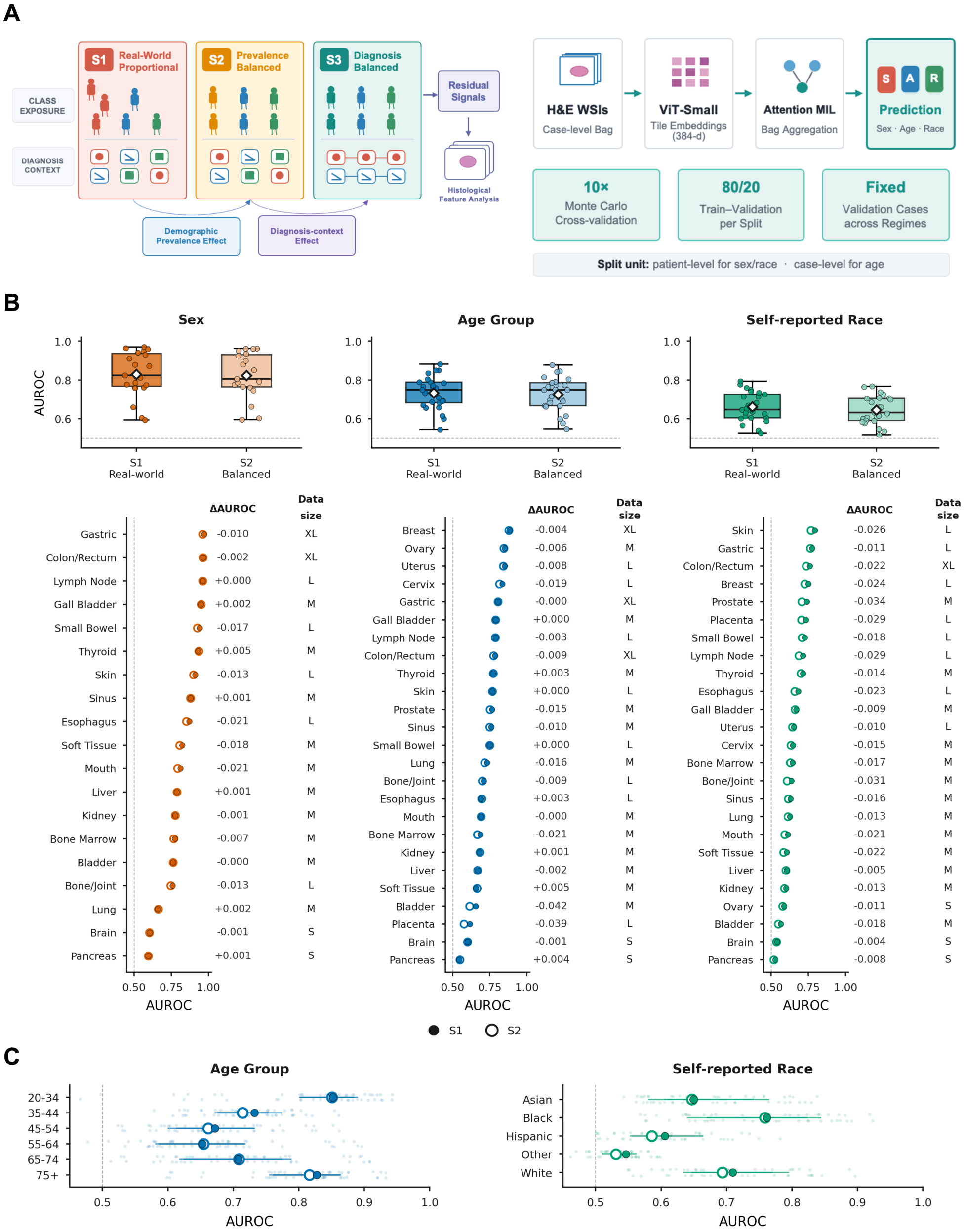}
  \caption{%
    \textbf{Experimental framework and pan-organ demographic signal across organs.}
    \textbf{(A) Experimental framework.}~S1 (observed-distribution), S2
    (demographic-balanced), and S3 (diagnosis-context grouped) case-level
    prediction: organ-matched H\&E slides, quality control, 20$\times$ tiling,
    in-house ViT-Small (SP22M) features, and gated multi-head AB-MIL.
    \textbf{(B) Pan-organ performance.}~S1/S2 AUROC distributions (top) and
    paired organ comparisons (bottom), with mean $\Delta$AUROC (S2$-$S1)
    across 10 matched MCCV folds and size tiers. Tiers use mean
    S1 training-case counts across folds per organ and target: XS, $<500$;
    S, $500$--$<2{,}000$; M, $2{,}000$--$<10{,}000$;
    L, $10{,}000$--$<30{,}000$; XL, $\geq30{,}000$.
    \textbf{(C) Subgroup performance.}~Age-group and self-reported-race AUROC
    under S1/S2. Large markers and horizontal bars denote cross-organ medians
    and interquartile ranges; background points show individual organs.
  }
  \label{fig:results}
\end{figure}

Demographic predictability under observed-distribution training (S1) was
widespread but varied across organs and targets. Across 69 organ--demographic
combinations (25 age-group, 25 self-reported-race, and 19 sex tasks), each
evaluated over ten MCCV splits, sex had the highest pan-organ mean AUROC
$\pm$ inter-organ SD:
$0.828 \pm 0.119$ (range, $0.594$--$0.970$). Age group followed at
$0.733 \pm 0.084$ (range, $0.545$--$0.882$), then self-reported race at
$0.662 \pm 0.074$ (range, $0.526$--$0.793$; Fig.~\ref{fig:results}B;
Supplementary Table~\ref{tab:supp-full-ladder}).
The \href{https://sinai-computational-pathology.github.io/saif-results-atlas/}{interactive Pan-organ Results Atlas}
provides a companion visualization of aggregate organ- and demographic-specific
results across training controls.

Organ-level performance differed within as well as across the subspecialty
groups in Fig.~\ref{fig:framework}C. Across the four gastrointestinal organs,
sex AUROC was highest in gastric tissue ($0.970$) and colon/rectum ($0.965$),
followed by small bowel ($0.942$) and esophagus ($0.872$). Breast had the
highest age-group AUROC ($0.882$); all three gynecologic organs also had
high age-group performance (ovary and uterus, $0.850$; cervix, $0.834$),
but lower self-reported-race performance ($0.588$--$0.653$).
Self-reported-race AUROC was highest in skin ($0.793$), with gastric tissue
and colon/rectum also among the highest-performing organs ($0.776$ and
$0.759$). Pancreas had the lowest AUROC for all three targets: $0.594$ for
sex, $0.545$ for age group, and $0.526$ for self-reported race
(Supplementary Table~\ref{tab:supp-full-ladder}).

\mainheading{Demographic predictability under class balancing}
\label{sec:results-s1s2}

Compared with S1, demographic-balanced training (S2) produced small
reductions in pan-organ mean AUROC: $0.006$ for sex ($0.822$ versus $0.828$),
$0.007$ for age group ($0.726$ versus $0.733$), and $0.018$ for self-reported
race ($0.644$ versus $0.662$; Fig.~\ref{fig:results}B; Supplementary
Table~\ref{tab:supp-full-ladder}).
The reduction was most consistent for self-reported race: all 25 organs
showed lower AUROC (median difference, $-0.017$; range, $-0.034$ to $-0.004$),
and 16 were lower in all ten matched folds (Fig.~\ref{fig:results}B).
Age and sex showed smaller, more heterogeneous changes, with median
differences of $-0.003$ and $-0.001$ and lower AUROC in 17 of 25 and
12 of 19 organs, respectively. Exploratory paired tests and BH-adjusted
results are reported in Supplementary Table~\ref{tab:supp-regime-statistics}.

Age-group AUROC followed an approximately U-shaped pattern under both S1 and
S2, with higher performance in the youngest and oldest groups.
Self-reported-race performance was heterogeneous, with Black or African
American and White generally higher and Other near chance; S2 largely
preserved this subgroup ordering despite modest overall attenuation.
Performance did not simply track training-set abundance: the largest age
groups had lower AUROC than the less frequent age extremes, and White had more
training cases but lower AUROC than Black or African American
(Fig.~\ref{fig:results}C; Supplementary Table~\ref{tab:subgroup_auroc}).

\begin{figure}[p]
  \centering
  \includegraphics[width=\textwidth]{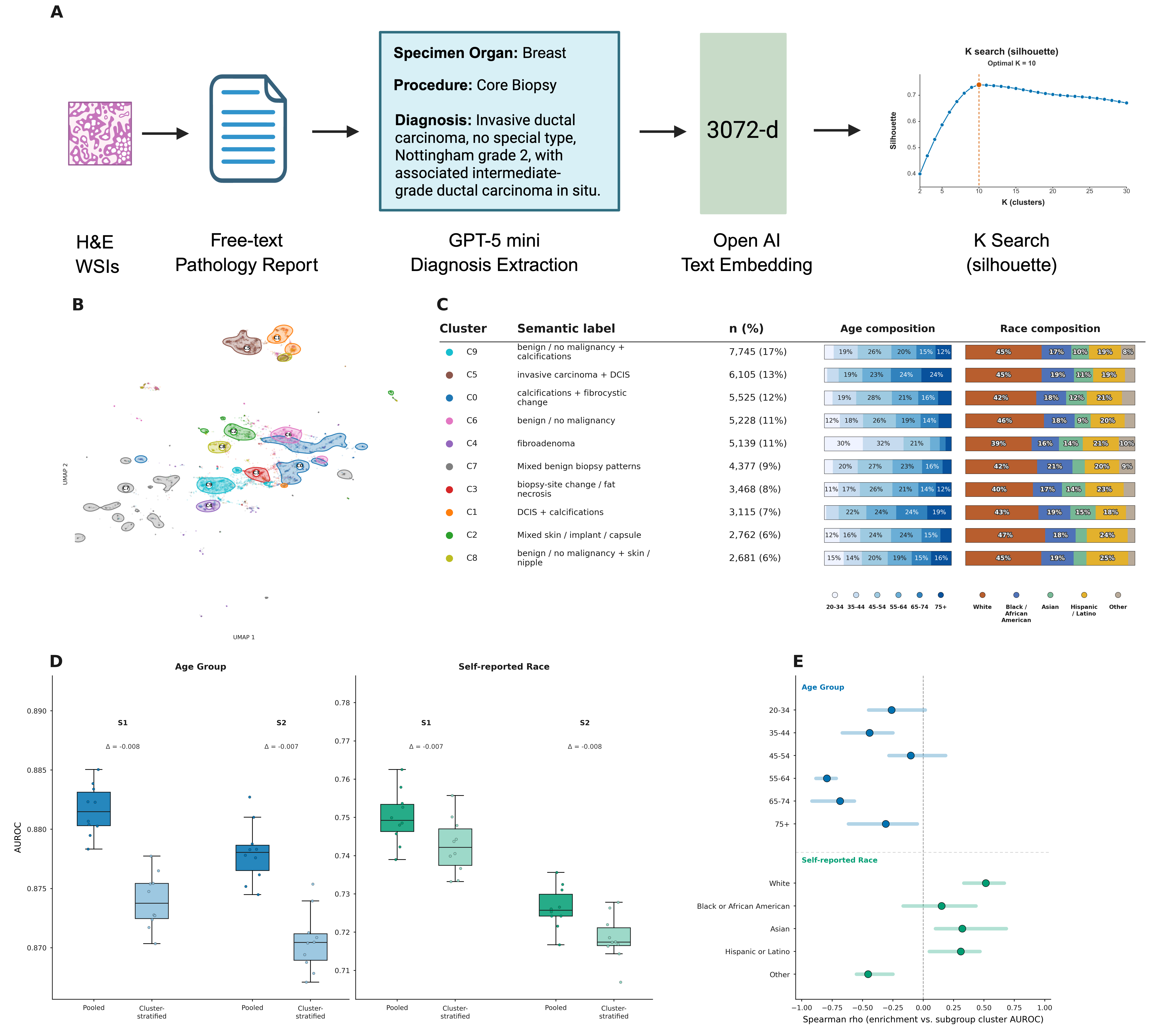}
  \caption{%
    \textbf{Breast as a worked example of diagnosis-context stratification.}
    \textbf{(A) Diagnosis-context pipeline.}~Free-text pathology reports were
    parsed into specimen-level diagnoses, encoded as 3,072-dimensional text
    embeddings, and grouped by $k$-means clustering; the number of clusters
    was selected by silhouette analysis.
    \textbf{(B) Diagnosis-cluster map.}~Uniform manifold approximation and projection (UMAP) of the resulting
    $K=10$ breast diagnosis clusters, shown by kernel-density contours.
    \textbf{(C) Cluster composition.}~Cluster size, semantic label, and
    age-group and self-reported-race composition across 46,145 unique breast
    cases, ranked by cluster size; age and race proportions use 45,750 and
    29,343 known-label cases, respectively. Labels summarize diagnosis-text
    patterns and were not independently validated by pathologists. Panels B--E use canonical $K=10$ assignments.
    \textbf{(D) Cluster-stratified performance.}~Pooled validation AUROC
    versus case-count-weighted within-cluster (cluster-stratified) AUROC for
    age group and self-reported race under S1 and S2. Boxes summarize 10 MCCV
    folds; annotations show the stratified-minus-pooled difference
    ($\Delta$).
    \textbf{(E) Enrichment--performance association.}~Spearman correlation
    between subgroup enrichment (percentage points relative to the eligible
    breast cohort) and within-cluster subgroup AUROC under S1; horizontal
    bars show leave-one-cluster-out sensitivity ranges.
  }
  \label{fig:stratification}
\end{figure}

\mainheading{Diagnosis context and within-cluster predictability}
\label{sec:results-strat}

Because diagnosis composition may contribute to demographic predictability,
we examined whether demographic-associated information remained detectable
when evaluation was restricted to cases with similar diagnostic contexts.
Breast served as a worked example for examining demographic composition
and predictability within diagnosis contexts. Diagnosis-text embeddings were
grouped into ten clusters linked to 46,145 unique cases
(Fig.~\ref{fig:stratification}A--C). Diagnosis-text patterns included fibrocystic
changes with calcifications, invasive carcinoma, fibroadenoma, and benign
parenchyma.
Demographic composition varied across clusters. Ages 20--34 accounted for
$30.3\%$ of cases with known age in the fibroadenoma cluster, compared with
$10.2\%$ across the breast cohort, an enrichment of $20.1$ percentage points.
In the invasive-carcinoma cluster, ages $\geq75$ accounted for $23.6\%$
versus $12.5\%$ overall (Fig.~\ref{fig:stratification}C). Associations between
diagnosis-cluster membership and demographic composition were also observed
across organs (Supplementary Table~\ref{tab:supp-cluster-quality}); cluster-level
demographic frequencies can also be explored in the
\href{https://sinai-computational-pathology.github.io/saif-results-atlas/}{interactive Pan-organ Results Atlas}.

Within-cluster evaluation produced modest attenuation of breast demographic
predictability. AUROC was computed within diagnosis clusters and aggregated
using case-count weighting for comparison with pooled validation AUROC.
Under S1, age-group AUROC decreased from $0.882$ pooled to $0.874$
cluster-stratified, and self-reported-race AUROC from $0.750$ to $0.743$,
with attenuation of $0.007$--$0.008$ under both S1 and S2
(Fig.~\ref{fig:stratification}D). Associations between subgroup enrichment
and within-cluster AUROC were negative for all six age groups and mixed in
sign for self-reported race (Fig.~\ref{fig:stratification}E).

Across all 69 organ--demographic combinations, within-cluster evaluation
using the primary silhouette-selected $K$ produced mean AUROC reductions
of $0.0216$ for age group, $0.0121$ for self-reported race, and $0.0095$
for sex under S1, and $0.0190$, $0.0118$, and $0.0098$, respectively,
under S2. The evaluation-only random-label control remained within $0.001$
of zero in aggregate (Supplementary
Tables~\ref{tab:supp-a4-aggregate}--\ref{tab:supp-a4-full}; Supplementary
Fig.~\ref{sfig:panorgan-diagnosis-stratified}).

\mainheading{Predictability under diagnosis-context grouped training}
\label{sec:results-grouped-training}

\begin{figure}[p]
  \centering
  \includegraphics[width=\textwidth,height=0.71\textheight,keepaspectratio]{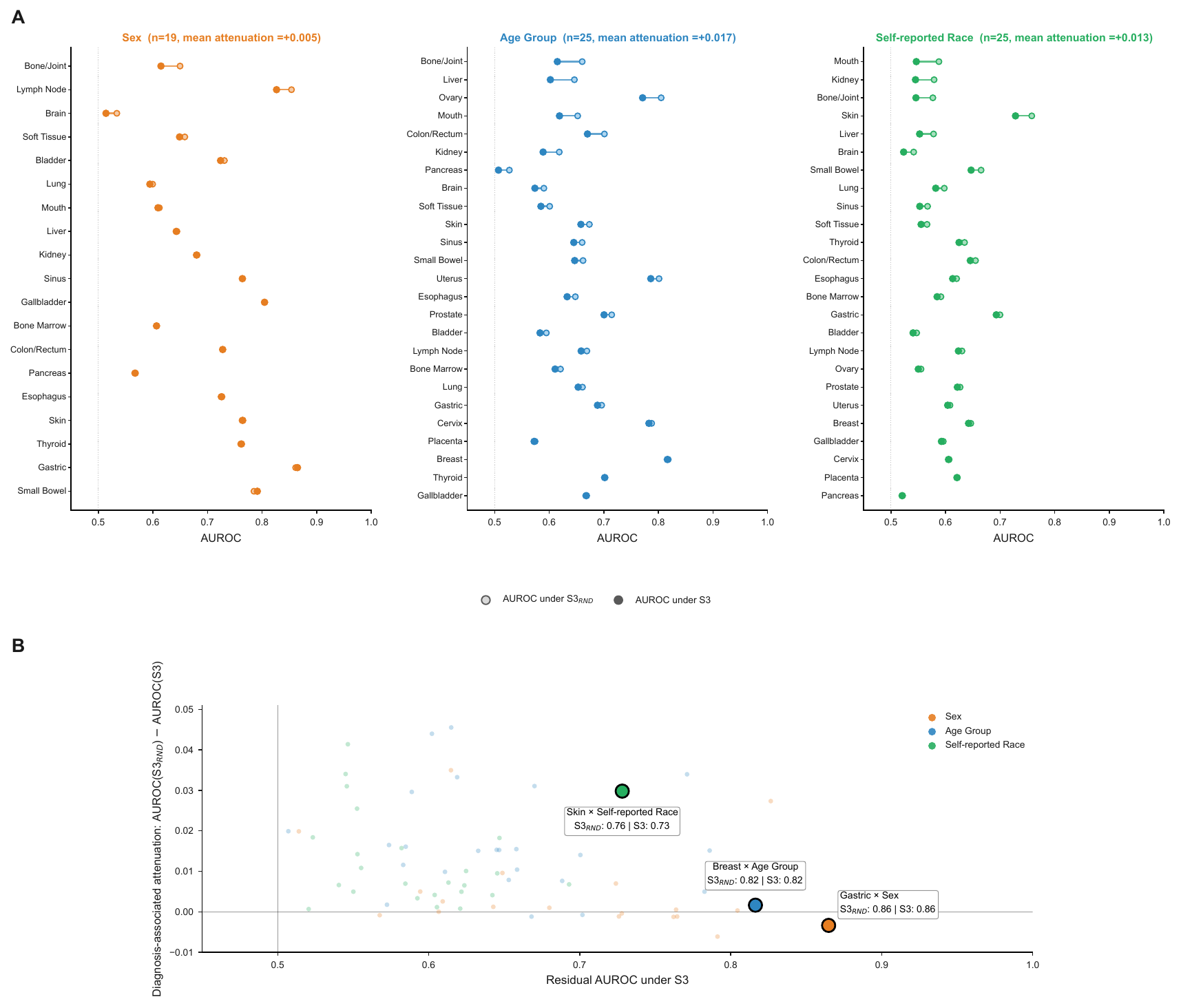}
  \caption{%
    \textbf{Residual demographic predictability under diagnosis-context grouped training.}
    \textbf{(A)}~Paired organ-level S3$_\text{RND}$ and S3 AUROC,
    ranked by the signed AUROC difference (19 sex, 25 age-group, and
    25 self-reported-race tasks). Prostate age uses four common classes and
    1,954 matched validation cases per fold; placenta age uses three classes
    and 2,872 matched cases. All S3 comparisons use epoch-50 evaluation
    (\suppmethodref{supp:class-comparability}). Markers distinguish the two training regimes; exploratory paired tests
    and BH-adjusted results are reported in Supplementary
    Table~\ref{tab:supp-regime-statistics}.
    \textbf{(B)}~Residual S3 AUROC versus the difference,
    AUROC(S3$_\text{RND}$)$-$AUROC(S3), across 69 combinations;
    positive values indicate lower S3 AUROC.
    Prostate age uses the harmonized S3 AUROC and difference;
    all other points retain their original evaluation definitions.
  }
  \label{fig:s3}
\end{figure}

Cluster-constrained grouped training (S3) yielded modestly
lower pan-organ AUROC than grouped demographic balancing without diagnosis
constraints (S3$_\text{RND}$). We define the signed AUROC difference as
$\Delta$AUROC $=$ AUROC(S3$_\text{RND}$) $-$ AUROC(S3), so
positive values indicate lower performance under cluster-constrained training.
Mean differences were $+0.005$ for sex, $+0.017$ for age group, and $+0.013$
for self-reported race (Fig.~\ref{fig:s3}A; Supplementary
Table~\ref{tab:supp-s3-increment}). Residual S3 pan-organ mean
AUROCs were $0.697$, $0.653$, and $0.594$, respectively, using the
matched S3 evaluation population for age (Fig.~\ref{fig:s3}; original
regime-specific values are retained in Supplementary Table~\ref{tab:supp-full-ladder}). The prostate age comparison used four
common classes and matched validation cases, with an AUROC difference of
$0.014$ (\suppmethodref{supp:class-comparability}).

Relative to S2, S3$_\text{RND}$ had lower mean AUROC for sex
($0.702$ versus $0.822$, a difference of $-0.121$) and self-reported race
($0.607$ versus $0.644$, a difference of $-0.037$); the corresponding
S3 differences were $-0.126$ and $-0.050$. These comparisons
reflect broader training-regime differences. Pan-organ S2/S3 age comparisons were omitted because evaluation class sets
differed in prostate and placenta (\suppmethodref{supp:class-comparability}).

Residual predictability and its sensitivity to diagnosis constraints varied
across the 69 combinations with comparable S3 evaluations
(Fig.~\ref{fig:s3}B). Sex prediction in gastric tissue retained an AUROC of
$0.865$ with little difference between regimes ($-0.003$), whereas sex
prediction in lymph nodes retained $0.827$ with a larger difference
($+0.027$). Age-group prediction in breast and prediction of self-reported
race in skin retained AUROCs of $0.816$ and $0.728$, with differences of
$+0.002$ and $+0.030$, respectively.

\FloatBarrier

\mainheading{Literature-guided hypotheses for selected signals}
\label{sec:results-hypotheses}

We developed the public
\href{https://sinai-computational-pathology.github.io/saif-community-hub/}{SAIF Hypothesis Explorer}
to organize candidate explanations and supporting, contradictory, and limiting
literature, and to facilitate further investigation and community discussion.
Candidates were generated through targeted LLM-assisted literature searching
followed by source checking (\headingref{sec:methods-hypothesis-map}).
Sex prediction in gastric tissue, age-group prediction in breast, and prediction
of self-reported race in skin received detailed attention because of their
substantial residual performance and value as representative settings for
further histomorphological investigation.

For sex prediction in gastric tissue, selected gastric cancer cohorts support
sex differences in subtype composition\cite{lu2016signetring,choi2022gastric}, while studies of
\textit{Helicobacter pylori}-associated and benign gastric disease suggest
context-dependent mucosal differences related to infection, subsite, and
eradication history\cite{kodama2021mucosal,kato2004mucosal}. These findings identify
cancer subtype composition and mucosal disease context as candidate
explanations for the retained signal.

For age-group prediction in breast, retained performance motivated review
of age-associated tissue changes. Terminal duct lobular unit (TDLU)
involution\cite{figueroa2014tdlu,radisky2016tdlu} and H\&E
morphometry\cite{sandhu2016morphometry} support differences in epithelial
cellularity and stromal/adipose composition as candidate correlates.
Reported associations between parity and lobular measures indicate that such
features are not specific to chronological age\cite{ogony2022parous}. These
observations are cross-sectional and do not establish that the model uses
these features.

For prediction of self-reported race in skin, substantial retained
predictability motivated investigation of both tissue and sampling context.
Epidermal pigmentation emerged as a candidate tissue correlate, supported
by studies of epidermal melanin and melanosome
distribution\cite{thong2003melanosome,alaluf2002melanin}. Some supporting
evidence used modalities or population definitions that do not map directly
to routine H\&E or the categories used here; anatomical-site and exposure
differences remain contextual alternatives, and measured ultraviolet
sensitivity has been reported to predict photodamage better than
racial or ethnic origin\cite{tadokoro2003uv}. Differential biopsy practice
is a further contextual candidate: within psoriasis, biopsy frequency has
been reported to differ by race and ethnicity\cite{ahmed2024psoriasis}, and
selected psoriasis biopsy series document a broad histological
spectrum\cite{chau2017psoriasis}, so diagnosis- and indication-specific
sampling could alter the lesion mix across label groups.

\FloatBarrier

\section{Discussion}
\label{sec:discussion}

This study systematically mapped demographic-associated information recoverable from routine clinical histopathology across 25 organs and 69 organ--demographic combinations. Using deep learning models as analytical tools, we found that information associated with sex, age group, and self-reported race was broadly detectable, although its recoverability varied substantially across organs and demographic attributes. Demographic-class balancing produced relatively small changes in model performance, and additional diagnosis-context analyses showed modest attenuation while leaving substantial signals in multiple tissues. Together, these findings establish a pan-organ reference for investigating demographic-associated variation in histopathology and its implications for computational pathology research.

Demographic-associated information in clinical histopathology may reflect differences in demographic representation and diagnostic composition, in addition to tissue-associated characteristics. To examine the sensitivity of the observed signals to these factors, we compared observed-distribution training (S1) with demographic-balanced training (S2), which produced relatively small changes in performance. Evaluation within diagnosis-text clusters further showed modest attenuation relative to pooled evaluation. Under grouped training, introducing diagnosis-context constraints (S3) yielded additional, generally modest attenuation compared with random grouping without these constraints (S3$_\text{RND}$). Notably, substantial demographic-associated signals remained across multiple organs under the evaluated controls, motivating further investigation of tissue-associated information and clinical or contextual factors not captured by these analyses.

The heterogeneity of demographic-associated information across organs raises questions about the histological and contextual features underlying these signals. Such information may be localized to specific tissue compartments, reflected in cellular morphology, or distributed across broader tissue regions. To facilitate further investigation, we selected three representative settings---sex in gastric tissue, age group in breast, and self-reported race in skin---and used literature-guided searching to identify plausible histomorphological and clinical correlates. Our previous dermatopathology study, in which attention and ablation analyses highlighted the epidermis as an important region for self-reported-race classification, illustrates how spatial analyses can help investigate candidate sources of information\cite{chen2025saif_skin}. The hypotheses assembled here provide starting points for targeted pathology review, quantitative feature measurement, and diagnosis-matched analyses, while inviting further investigation of alternative explanations and previously unrecognized tissue-associated patterns.

The widespread availability of demographic-associated information in histopathology has implications for computational pathology model development and clinical translation. Such information may provide useful context for certain clinical tasks, but it may also become a source of shortcut learning when models exploit demographic correlations that do not generalize across populations or care settings\cite{ongly2024shortcut,jones2024causal}. Whether these associations are informative or problematic depends on the intended task, the model's use of the information, and the population in which it is applied. Our findings therefore support more systematic characterization of patient populations and demographic subgroup performance during model development and evaluation\cite{vaidya2024demographic,lin2025fairpath}. Where data permit, studies should report performance across relevant demographic groups, including self-reported race, sex, and age. At a minimum, clear reporting of cohort demographic composition and population restrictions can help researchers assess the applicability of a model to new clinical settings. The pan-organ map presented here may help prioritize organ-specific investigations, subgroup validation, and assessments of potential demographic shortcuts in downstream pathology models.

Several considerations remain important for interpreting these findings. Self-reported race may serve as an imperfect proxy for genetic ancestry while also reflecting social, environmental, and healthcare-related factors. Future studies incorporating ancestry-informed measures could provide complementary perspectives on the sources of demographic-associated information in histopathology. The single-health-system cohort provided a relatively consistent clinical and imaging environment, although independent cohorts will be important for evaluating the reproducibility of organ-specific patterns. Diagnosis-text clustering captured broad diagnostic contexts but could not fully account for disease heterogeneity, clinical history, sampling practices, or other potential confounders. The present study also did not directly localize the histological features underlying the observed signals. Future investigations combining more granular clinical stratification with tissue-region, tile-level, and cellular analyses may help identify the morphological and contextual sources of demographic-associated information. By providing a systematic map and literature-informed candidate hypotheses, this study offers a foundation for further investigation of these signals and their relevance to computational pathology.

\section*{Ethics approval}
This retrospective study was approved by the Institutional Review Board at the
Icahn School of Medicine at Mount Sinai (Protocol STUDY-20-00338). Informed consent
was waived under the approved protocol.

\section*{Data availability}
Two complementary interactive resources accompany the manuscript:
the \href{https://sinai-computational-pathology.github.io/saif-results-atlas/}{Pan-organ Results Atlas}
for aggregate organ- and demographic-specific results, and the
\href{https://sinai-computational-pathology.github.io/saif-community-hub/}{SAIF Hypothesis Explorer}
for candidate explanations, supporting and limiting literature, and community
discussion. The hypotheses are proposed explanations, not validated findings.

\bibliographystyle{naturemag}
\bibliography{references}

\begin{thebibliography}{10}
\expandafter\ifx\csname url\endcsname\relax
  \def\url#1{\texttt{#1}}\fi
\expandafter\ifx\csname urlprefix\endcsname\relax\def\urlprefix{URL }\fi
\providecommand{\bibinfo}[2]{#2}
\providecommand{\eprint}[2][]{\url{#2}}

\bibitem{barr1949nuclear}
\bibinfo{author}{Barr, M.~L.} \& \bibinfo{author}{Bertram, E.~G.}
\newblock \bibinfo{title}{A morphological distinction between neurones of the
  male and female, and the behaviour of the nucleolar satellite during
  accelerated nucleoprotein synthesis}.
\newblock \emph{\bibinfo{journal}{Nature}} \textbf{\bibinfo{volume}{163}},
  \bibinfo{pages}{676--677} (\bibinfo{year}{1949}).

\bibitem{davidson1954drumstick}
\bibinfo{author}{Davidson, W.~M.} \& \bibinfo{author}{Smith, D.~R.}
\newblock \bibinfo{title}{A morphological sex difference in the
  polymorphonuclear neutrophil leucocytes}.
\newblock \emph{\bibinfo{journal}{BMJ}} \textbf{\bibinfo{volume}{2}},
  \bibinfo{pages}{6--7} (\bibinfo{year}{1954}).

\bibitem{gerstein2025imageqtl}
\bibinfo{author}{Meng, R.} \emph{et~al.}
\newblock \bibinfo{title}{Machine-learning models based on histological images
  from healthy donors identify {imageQTLs} and predict chronological age}.
\newblock \emph{\bibinfo{journal}{Proc. Natl Acad. Sci. USA}}
  \textbf{\bibinfo{volume}{122}}, \bibinfo{pages}{e2423469122}
  (\bibinfo{year}{2025}).

\bibitem{braun2023histopathologic}
\bibinfo{author}{Marx, G.~A.} \emph{et~al.}
\newblock \bibinfo{title}{Histopathologic brain age estimation via multiple
  instance learning}.
\newblock \emph{\bibinfo{journal}{Acta Neuropathol.}}
  \textbf{\bibinfo{volume}{146}}, \bibinfo{pages}{785--802}
  (\bibinfo{year}{2023}).

\bibitem{yadav2026pathstar}
\bibinfo{author}{Yadav, A.}, \bibinfo{author}{Alvarez, K.},
  \bibinfo{author}{Chechenina, A.} \emph{et~al.}
\newblock \bibinfo{title}{Mapping structural aging across human tissues reveals
  tissue-specific trajectories and coordinated deterioration}.
\newblock \emph{\bibinfo{journal}{Nat. Aging}} \textbf{\bibinfo{volume}{6}},
  \bibinfo{pages}{2013--2027} (\bibinfo{year}{2026}).

\bibitem{williams2023skin}
\bibinfo{author}{Williams, K.~A.}, \bibinfo{author}{Wondimu, B.},
  \bibinfo{author}{Ajayi, A.~M.} \& \bibinfo{author}{Sokumbi, O.}
\newblock \bibinfo{title}{Skin of color in dermatopathology: does color
  matter?}
\newblock \emph{\bibinfo{journal}{Hum. Pathol.}}
  \textbf{\bibinfo{volume}{140}}, \bibinfo{pages}{240--266}
  (\bibinfo{year}{2023}).

\bibitem{bigler2011prostate}
\bibinfo{author}{Bigler, S.~A.}, \bibinfo{author}{Pound, C.~R.} \&
  \bibinfo{author}{Zhou, X.}
\newblock \bibinfo{title}{A retrospective study on pathologic features and
  racial disparities in prostate cancer}.
\newblock \emph{\bibinfo{journal}{Prostate Cancer}}
  \textbf{\bibinfo{volume}{2011}}, \bibinfo{pages}{1--7}
  (\bibinfo{year}{2011}).

\bibitem{awasthi2021prostate}
\bibinfo{author}{Awasthi, S.} \emph{et~al.}
\newblock \bibinfo{title}{Comparative genomics reveals distinct
  immune-oncologic pathways in {African American} men with prostate cancer}.
\newblock \emph{\bibinfo{journal}{Clin. Cancer Res.}}
  \textbf{\bibinfo{volume}{27}}, \bibinfo{pages}{320--329}
  (\bibinfo{year}{2021}).

\bibitem{campanella2019clinical}
\bibinfo{author}{Campanella, G.} \emph{et~al.}
\newblock \bibinfo{title}{Clinical-grade computational pathology using weakly
  supervised deep learning on whole slide images}.
\newblock \emph{\bibinfo{journal}{Nat. Med.}} \textbf{\bibinfo{volume}{25}},
  \bibinfo{pages}{1301--1309} (\bibinfo{year}{2019}).

\bibitem{bulten2022panda}
\bibinfo{author}{Bulten, W.} \emph{et~al.}
\newblock \bibinfo{title}{Artificial intelligence for diagnosis and {Gleason}
  grading of prostate cancer: the {PANDA} challenge}.
\newblock \emph{\bibinfo{journal}{Nat. Med.}} \textbf{\bibinfo{volume}{28}},
  \bibinfo{pages}{154--163} (\bibinfo{year}{2022}).

\bibitem{kather2019msi}
\bibinfo{author}{Kather, J.~N.} \emph{et~al.}
\newblock \bibinfo{title}{Deep learning can predict microsatellite instability
  directly from histology in gastrointestinal cancer}.
\newblock \emph{\bibinfo{journal}{Nat. Med.}} \textbf{\bibinfo{volume}{25}},
  \bibinfo{pages}{1054--1056} (\bibinfo{year}{2019}).

\bibitem{campanella2025eagle}
\bibinfo{author}{Campanella, G.} \emph{et~al.}
\newblock \bibinfo{title}{Real-world deployment of a fine-tuned pathology
  foundation model for lung cancer biomarker detection}.
\newblock \emph{\bibinfo{journal}{Nat. Med.}} \textbf{\bibinfo{volume}{31}},
  \bibinfo{pages}{3002--3010} (\bibinfo{year}{2025}).

\bibitem{volinsky2024hector}
\bibinfo{author}{Volinsky-Fremond, S.} \emph{et~al.}
\newblock \bibinfo{title}{Prediction of recurrence risk in endometrial cancer
  with multimodal deep learning}.
\newblock \emph{\bibinfo{journal}{Nat. Med.}} \textbf{\bibinfo{volume}{30}},
  \bibinfo{pages}{1962--1973} (\bibinfo{year}{2024}).

\bibitem{graham2019hovernet}
\bibinfo{author}{Graham, S.} \emph{et~al.}
\newblock \bibinfo{title}{{Hover-Net}: Simultaneous segmentation and
  classification of nuclei in multi-tissue histology images}.
\newblock \emph{\bibinfo{journal}{Med. Image Anal.}}
  \textbf{\bibinfo{volume}{58}}, \bibinfo{pages}{101563}
  (\bibinfo{year}{2019}).

\bibitem{horst2024cellvit}
\bibinfo{author}{H{\"o}rst, F.} \emph{et~al.}
\newblock \bibinfo{title}{{CellViT}: Vision transformers for precise cell
  segmentation and classification}.
\newblock \emph{\bibinfo{journal}{Med. Image Anal.}}
  \textbf{\bibinfo{volume}{94}}, \bibinfo{pages}{103143}
  (\bibinfo{year}{2024}).

\bibitem{chen2024uni}
\bibinfo{author}{Chen, R.~J.} \emph{et~al.}
\newblock \bibinfo{title}{Towards a general-purpose foundation model for
  computational pathology}.
\newblock \emph{\bibinfo{journal}{Nat. Med.}} \textbf{\bibinfo{volume}{30}},
  \bibinfo{pages}{850--862} (\bibinfo{year}{2024}).

\bibitem{xu2024gigapath}
\bibinfo{author}{Xu, H.} \emph{et~al.}
\newblock \bibinfo{title}{A whole-slide foundation model for digital pathology
  from real-world data}.
\newblock \emph{\bibinfo{journal}{Nature}} \textbf{\bibinfo{volume}{630}},
  \bibinfo{pages}{181--188} (\bibinfo{year}{2024}).

\bibitem{vorontsov2024virchow}
\bibinfo{author}{Vorontsov, E.} \emph{et~al.}
\newblock \bibinfo{title}{A foundation model for clinical-grade computational
  pathology and rare cancers detection}.
\newblock \emph{\bibinfo{journal}{Nat. Med.}} \textbf{\bibinfo{volume}{30}},
  \bibinfo{pages}{2924--2935} (\bibinfo{year}{2024}).

\bibitem{ding2025titan}
\bibinfo{author}{Ding, T.} \emph{et~al.}
\newblock \bibinfo{title}{A multimodal whole-slide foundation model for
  pathology}.
\newblock \emph{\bibinfo{journal}{Nat. Med.}} \textbf{\bibinfo{volume}{31}},
  \bibinfo{pages}{3749--3761} (\bibinfo{year}{2025}).

\bibitem{lu2023conch}
\bibinfo{author}{Lu, M.~Y.} \emph{et~al.}
\newblock \bibinfo{title}{A visual-language foundation model for computational
  pathology}.
\newblock \emph{\bibinfo{journal}{Nat. Med.}} \textbf{\bibinfo{volume}{30}},
  \bibinfo{pages}{863--874} (\bibinfo{year}{2024}).

\bibitem{campanella2025benchmark}
\bibinfo{author}{Campanella, G.} \emph{et~al.}
\newblock \bibinfo{title}{A clinical benchmark of public self-supervised
  pathology foundation models}.
\newblock \emph{\bibinfo{journal}{Nat. Commun.}} \textbf{\bibinfo{volume}{16}},
  \bibinfo{pages}{3640} (\bibinfo{year}{2025}).

\bibitem{neidlinger2025benchmark}
\bibinfo{author}{Neidlinger, P.} \emph{et~al.}
\newblock \bibinfo{title}{Benchmarking foundation models as feature extractors
  for weakly supervised computational pathology}.
\newblock \emph{\bibinfo{journal}{Nat. Biomed. Eng.}}
  \textbf{\bibinfo{volume}{10}}, \bibinfo{pages}{1113--1123}
  (\bibinfo{year}{2026}).

\bibitem{adleberg2022predicting}
\bibinfo{author}{Adleberg, J.} \emph{et~al.}
\newblock \bibinfo{title}{Predicting patient demographics from chest
  radiographs with deep learning}.
\newblock \emph{\bibinfo{journal}{J. Am. Coll. Radiol.}}
  \textbf{\bibinfo{volume}{19}}, \bibinfo{pages}{1151--1161}
  (\bibinfo{year}{2022}).

\bibitem{gichoya2022ai}
\bibinfo{author}{Gichoya, J.~W.} \emph{et~al.}
\newblock \bibinfo{title}{{AI} recognition of patient race in medical imaging:
  a modelling study}.
\newblock \emph{\bibinfo{journal}{Lancet Digit. Health}}
  \textbf{\bibinfo{volume}{4}}, \bibinfo{pages}{e406--e414}
  (\bibinfo{year}{2022}).

\bibitem{joo2025vlfoundation}
\bibinfo{author}{Yang, Y.} \emph{et~al.}
\newblock \bibinfo{title}{Demographic bias of expert-level vision--language
  foundation models in medical imaging}.
\newblock \emph{\bibinfo{journal}{Sci. Adv.}} \textbf{\bibinfo{volume}{11}},
  \bibinfo{pages}{eadq0305} (\bibinfo{year}{2025}).

\bibitem{vaidya2024demographic}
\bibinfo{author}{Vaidya, A.} \emph{et~al.}
\newblock \bibinfo{title}{Demographic bias in misdiagnosis by computational
  pathology models}.
\newblock \emph{\bibinfo{journal}{Nat. Med.}} \textbf{\bibinfo{volume}{30}},
  \bibinfo{pages}{1174--1190} (\bibinfo{year}{2024}).

\bibitem{chen2025saif_skin}
\bibinfo{author}{Chen, S.} \emph{et~al.}
\newblock \bibinfo{title}{Predicting patient self-reported race from skin
  histological images with deep learning}.
\newblock In \emph{\bibinfo{booktitle}{Workshop on Fairness of AI in Medical
  Imaging (FAIMI), MICCAI}}, vol. \bibinfo{volume}{15976} of
  \emph{\bibinfo{series}{Lecture Notes in Computer Science}},
  \bibinfo{pages}{104--114} (\bibinfo{publisher}{Springer},
  \bibinfo{year}{2025}).

\bibitem{ongly2024shortcut}
\bibinfo{author}{Ong~Ly, C.} \emph{et~al.}
\newblock \bibinfo{title}{Shortcut learning in medical {AI} hinders
  generalization: method for estimating {AI} model generalization without
  external data}.
\newblock \emph{\bibinfo{journal}{npj Digit. Med.}}
  \textbf{\bibinfo{volume}{7}}, \bibinfo{pages}{124} (\bibinfo{year}{2024}).

\bibitem{jones2024causal}
\bibinfo{author}{Jones, C.} \emph{et~al.}
\newblock \bibinfo{title}{A causal perspective on dataset bias in machine
  learning for medical imaging}.
\newblock \emph{\bibinfo{journal}{Nat. Mach. Intell.}}
  \textbf{\bibinfo{volume}{6}}, \bibinfo{pages}{138--146}
  (\bibinfo{year}{2024}).

\bibitem{yang2024limits}
\bibinfo{author}{Yang, Y.}, \bibinfo{author}{Zhang, H.},
  \bibinfo{author}{Gichoya, J.~W.}, \bibinfo{author}{Katabi, D.} \&
  \bibinfo{author}{Ghassemi, M.}
\newblock \bibinfo{title}{The limits of fair medical imaging {AI} in real-world
  generalization}.
\newblock \emph{\bibinfo{journal}{Nat. Med.}} \textbf{\bibinfo{volume}{30}},
  \bibinfo{pages}{2838--2848} (\bibinfo{year}{2024}).

\bibitem{lin2025fairpath}
\bibinfo{author}{Lin, S.-Y.} \emph{et~al.}
\newblock \bibinfo{title}{Contrastive learning enhances fairness in pathology
  artificial intelligence systems}.
\newblock \emph{\bibinfo{journal}{Cell Rep. Med.}}
  \textbf{\bibinfo{volume}{6}}, \bibinfo{pages}{102527} (\bibinfo{year}{2025}).

\bibitem{howard2021impact}
\bibinfo{author}{Howard, F.~M.} \emph{et~al.}
\newblock \bibinfo{title}{The impact of site-specific digital histology
  signatures on deep learning model accuracy and bias}.
\newblock \emph{\bibinfo{journal}{Nat. Commun.}} \textbf{\bibinfo{volume}{12}},
  \bibinfo{pages}{4423} (\bibinfo{year}{2021}).

\bibitem{biased2023tcga}
\bibinfo{author}{Dehkharghanian, T.} \emph{et~al.}
\newblock \bibinfo{title}{Biased data, biased {AI}: deep networks predict the
  acquisition site of {TCGA} images}.
\newblock \emph{\bibinfo{journal}{Diagn. Pathol.}}
  \textbf{\bibinfo{volume}{18}}, \bibinfo{pages}{67} (\bibinfo{year}{2023}).

\bibitem{huang2025flex}
\bibinfo{author}{Huang, Y.} \emph{et~al.}
\newblock \bibinfo{title}{Knowledge-guided adaptation of pathology foundation
  models effectively improves cross-domain generalization and demographic
  fairness}.
\newblock \emph{\bibinfo{journal}{Nat. Commun.}} \textbf{\bibinfo{volume}{16}},
  \bibinfo{pages}{11485} (\bibinfo{year}{2025}).

\bibitem{oakden2020hidden}
\bibinfo{author}{Oakden-Rayner, L.} \emph{et~al.}
\newblock \bibinfo{title}{Hidden stratification causes clinically meaningful
  failures in machine learning for medical imaging}.
\newblock In \emph{\bibinfo{booktitle}{Proceedings of the ACM Conference on
  Health, Inference, and Learning}}, \bibinfo{pages}{151--159}
  (\bibinfo{year}{2020}).

\bibitem{openai2025gpt5systemcard}
\bibinfo{author}{{OpenAI}}.
\newblock \bibinfo{title}{{GPT-5} system card}.
\newblock \bibinfo{type}{Tech. Rep.}, \bibinfo{institution}{OpenAI}
  (\bibinfo{year}{2025}).
\newblock \urlprefix\url{https://cdn.openai.com/gpt-5-system-card.pdf}.
\newblock \bibinfo{note}{13 August 2025}.

\bibitem{weng2024grandqc}
\bibinfo{author}{Weng, Z.} \emph{et~al.}
\newblock \bibinfo{title}{{GrandQC}: A comprehensive solution to quality
  control problem in digital pathology}.
\newblock \emph{\bibinfo{journal}{Nat. Commun.}} \textbf{\bibinfo{volume}{15}},
  \bibinfo{pages}{10685} (\bibinfo{year}{2024}).

\bibitem{campanella2023sp22m}
\bibinfo{author}{Campanella, G.} \emph{et~al.}
\newblock \bibinfo{title}{Computational pathology at health system scale --
  self-supervised foundation models from three billion images}.
\newblock \emph{\bibinfo{journal}{arXiv preprint arXiv:2310.07033}}
  (\bibinfo{year}{2023}).

\bibitem{ilse2018attention}
\bibinfo{author}{Ilse, M.}, \bibinfo{author}{Tomczak, J.~M.} \&
  \bibinfo{author}{Welling, M.}
\newblock \bibinfo{title}{Attention-based deep multiple instance learning}.
\newblock In \emph{\bibinfo{booktitle}{Proceedings of the 35th International
  Conference on Machine Learning}}, vol.~\bibinfo{volume}{80} of
  \emph{\bibinfo{series}{Proceedings of Machine Learning Research}},
  \bibinfo{pages}{2127--2136} (\bibinfo{publisher}{PMLR},
  \bibinfo{year}{2018}).
\newblock \urlprefix\url{https://proceedings.mlr.press/v80/ilse18a.html}.

\bibitem{lu2021data}
\bibinfo{author}{Lu, M.~Y.} \emph{et~al.}
\newblock \bibinfo{title}{Data-efficient and weakly supervised computational
  pathology on whole-slide images}.
\newblock \emph{\bibinfo{journal}{Nat. Biomed. Eng.}}
  \textbf{\bibinfo{volume}{5}}, \bibinfo{pages}{555--570}
  (\bibinfo{year}{2021}).

\bibitem{acs2024_nyc}
\bibinfo{author}{{U.S. Census Bureau}}.
\newblock \bibinfo{title}{{Hispanic} or {Latino} origin by race, {New York
  City} ({American} {Community} {Survey} 2024 1-year estimates, table
  {B03002})} (\bibinfo{year}{2025}).
\newblock
  \urlprefix\url{https://data.census.gov/table/ACSDT1Y2024.B03002?g=160XX00US3651000}.
\newblock \bibinfo{note}{Accessed 2026-07-23}.

\bibitem{lu2016signetring}
\bibinfo{author}{Lu, M.}, \bibinfo{author}{Yang, Z.}, \bibinfo{author}{Feng,
  Q.} \emph{et~al.}
\newblock \bibinfo{title}{The characteristics and prognostic value of signet
  ring cell histology in gastric cancer: a retrospective cohort study of 2199
  consecutive patients}.
\newblock \emph{\bibinfo{journal}{Medicine (Baltimore)}}
  \textbf{\bibinfo{volume}{95}}, \bibinfo{pages}{e4052} (\bibinfo{year}{2016}).

\bibitem{choi2022gastric}
\bibinfo{author}{Choi, Y.} \emph{et~al.}
\newblock \bibinfo{title}{Sex-based differences in histology, staging, and
  prognosis among 2983 gastric cancer surgery patients}.
\newblock \emph{\bibinfo{journal}{World journal of gastroenterology}}
  \textbf{\bibinfo{volume}{28}}, \bibinfo{pages}{933--947}
  (\bibinfo{year}{2022}).

\bibitem{kodama2021mucosal}
\bibinfo{author}{Kodama, M.} \emph{et~al.}
\newblock \bibinfo{title}{Gastric mucosal changes, and sex differences therein,
  after helicobacter pylori eradication: A long-term prospective follow-up
  study}.
\newblock \emph{\bibinfo{journal}{Journal of gastroenterology and hepatology}}
  \textbf{\bibinfo{volume}{36}}, \bibinfo{pages}{2210--2216}
  (\bibinfo{year}{2021}).

\bibitem{kato2004mucosal}
\bibinfo{author}{Kato, S.} \emph{et~al.}
\newblock \bibinfo{title}{Sex differences in mucosal response to helicobacter
  pylori infection in the stomach and variations in interleukin-8, cox-2 and
  trefoil factor family 1 gene expression}.
\newblock \emph{\bibinfo{journal}{Alimentary pharmacology \& therapeutics}}
  \textbf{\bibinfo{volume}{20 Suppl 1}}, \bibinfo{pages}{17--24}
  (\bibinfo{year}{2004}).

\bibitem{figueroa2014tdlu}
\bibinfo{author}{Figueroa, J.~D.} \emph{et~al.}
\newblock \bibinfo{title}{Terminal duct lobular unit involution of the normal
  breast: implications for breast cancer etiology}.
\newblock \emph{\bibinfo{journal}{Journal of the National Cancer Institute}}
  \textbf{\bibinfo{volume}{106}} (\bibinfo{year}{2014}).

\bibitem{radisky2016tdlu}
\bibinfo{author}{Radisky, D.~C.}, \bibinfo{author}{Visscher, D.~W.},
  \bibinfo{author}{Frank, R.~D.} \emph{et~al.}
\newblock \bibinfo{title}{Natural history of age-related lobular involution and
  impact on breast cancer risk}.
\newblock \emph{\bibinfo{journal}{Breast Cancer Res. Treat.}}
  \textbf{\bibinfo{volume}{155}}, \bibinfo{pages}{423--430}
  (\bibinfo{year}{2016}).

\bibitem{sandhu2016morphometry}
\bibinfo{author}{Sandhu, R.}, \bibinfo{author}{Chollet-Hinton, L.},
  \bibinfo{author}{Kirk, E.~L.}, \bibinfo{author}{Midkiff, B.} \&
  \bibinfo{author}{Troester, M.~A.}
\newblock \bibinfo{title}{Digital histologic analysis reveals morphometric
  patterns of age-related involution in breast epithelium and stroma}.
\newblock \emph{\bibinfo{journal}{Human pathology}}
  \textbf{\bibinfo{volume}{48}}, \bibinfo{pages}{60--8} (\bibinfo{year}{2016}).

\bibitem{ogony2022parous}
\bibinfo{author}{Ogony, J.} \emph{et~al.}
\newblock \bibinfo{title}{Towards defining morphologic parameters of normal
  parous and nulliparous breast tissues by artificial intelligence}.
\newblock \emph{\bibinfo{journal}{Breast cancer research : BCR}}
  \textbf{\bibinfo{volume}{24}}, \bibinfo{pages}{45} (\bibinfo{year}{2022}).

\bibitem{thong2003melanosome}
\bibinfo{author}{Thong, H.~Y.}, \bibinfo{author}{Jee, S.~H.},
  \bibinfo{author}{Sun, C.~C.} \& \bibinfo{author}{Boissy, R.~E.}
\newblock \bibinfo{title}{The patterns of melanosome distribution in
  keratinocytes of human skin as one determining factor of skin colour}.
\newblock \emph{\bibinfo{journal}{The British journal of dermatology}}
  \textbf{\bibinfo{volume}{149}}, \bibinfo{pages}{498--505}
  (\bibinfo{year}{2003}).

\bibitem{alaluf2002melanin}
\bibinfo{author}{Alaluf, S.} \emph{et~al.}
\newblock \bibinfo{title}{Ethnic variation in melanin content and composition
  in photoexposed and photoprotected human skin}.
\newblock \emph{\bibinfo{journal}{Pigment cell research}}
  \textbf{\bibinfo{volume}{15}}, \bibinfo{pages}{112--8}
  (\bibinfo{year}{2002}).

\bibitem{tadokoro2003uv}
\bibinfo{author}{Tadokoro, T.} \emph{et~al.}
\newblock \bibinfo{title}{Uv-induced dna damage and melanin content in human
  skin differing in racial/ethnic origin}.
\newblock \emph{\bibinfo{journal}{FASEB journal : official publication of the
  Federation of American Societies for Experimental Biology}}
  \textbf{\bibinfo{volume}{17}}, \bibinfo{pages}{1177--9}
  (\bibinfo{year}{2003}).

\bibitem{ahmed2024psoriasis}
\bibinfo{author}{Ahmed, F.}, \bibinfo{author}{Fitzsimmons, R.},
  \bibinfo{author}{Chu, E.~Y.}, \bibinfo{author}{Shin, D.~B.} \&
  \bibinfo{author}{Takeshita, J.}
\newblock \bibinfo{title}{Frequency of skin biopsies for psoriasis by race and
  ethnicity}.
\newblock \emph{\bibinfo{journal}{JAMA dermatology}}
  \textbf{\bibinfo{volume}{160}}, \bibinfo{pages}{1003--1005}
  (\bibinfo{year}{2024}).

\bibitem{chau2017psoriasis}
\bibinfo{author}{Chau, T.} \emph{et~al.}
\newblock \bibinfo{title}{Psoriasis or not? review of 51 clinically confirmed
  cases reveals an expanded histopathologic spectrum of psoriasis}.
\newblock \emph{\bibinfo{journal}{Journal of cutaneous pathology}}
  \textbf{\bibinfo{volume}{44}}, \bibinfo{pages}{1018--1026}
  (\bibinfo{year}{2017}).

\bibitem{janssen1991gastric}
\bibinfo{author}{Janssen, C.~W.}, \bibinfo{author}{Maartmann-Moe, H.},
  \bibinfo{author}{Lie, R.~T.} \& \bibinfo{author}{Matre, R.}
\newblock \bibinfo{title}{Age and sex distribution of intestinal type and
  diffuse gastric carcinoma}.
\newblock \emph{\bibinfo{journal}{APMIS : acta pathologica, microbiologica, et
  immunologica Scandinavica}} \textbf{\bibinfo{volume}{99}},
  \bibinfo{pages}{78--82} (\bibinfo{year}{1991}).

\bibitem{derakhshan2009oesophageal}
\bibinfo{author}{Derakhshan, M.~H.} \emph{et~al.}
\newblock \bibinfo{title}{Oesophageal and gastric intestinal-type
  adenocarcinomas show the same male predominance due to a 17 year delayed
  development in females}.
\newblock \emph{\bibinfo{journal}{Gut}} \textbf{\bibinfo{volume}{58}},
  \bibinfo{pages}{16--23} (\bibinfo{year}{2009}).

\bibitem{kensler2020tdlu}
\bibinfo{author}{Kensler, K.~H.} \emph{et~al.}
\newblock \bibinfo{title}{Automated quantitative measures of terminal duct
  lobular unit involution and breast cancer risk}.
\newblock \emph{\bibinfo{journal}{Cancer epidemiology, biomarkers \&
  prevention}} \textbf{\bibinfo{volume}{29}}, \bibinfo{pages}{2358--2368}
  (\bibinfo{year}{2020}).

\bibitem{girardeau2009dermal}
\bibinfo{author}{Girardeau, S.}, \bibinfo{author}{Mine, S.},
  \bibinfo{author}{Pageon, H.} \& \bibinfo{author}{Asselineau, D.}
\newblock \bibinfo{title}{The caucasian and african skin types differ
  morphologically and functionally in their dermal component}.
\newblock \emph{\bibinfo{journal}{Experimental dermatology}}
  \textbf{\bibinfo{volume}{18}}, \bibinfo{pages}{704--11}
  (\bibinfo{year}{2009}).

\end{thebibliography}

\newpage
\appendix
\section*{Supplementary Material}

\renewcommand{\thetable}{S\arabic{table}}
\setcounter{table}{0}
\renewcommand{\thefigure}{S\arabic{figure}}
\setcounter{figure}{0}
\newcounter{suppnote}
\renewcommand{\thesuppnote}{S\arabic{suppnote}}

\subsection*{Supplementary Figures}

\begin{figure}[htbp]
  \centering
  \includegraphics[width=0.94\textwidth]{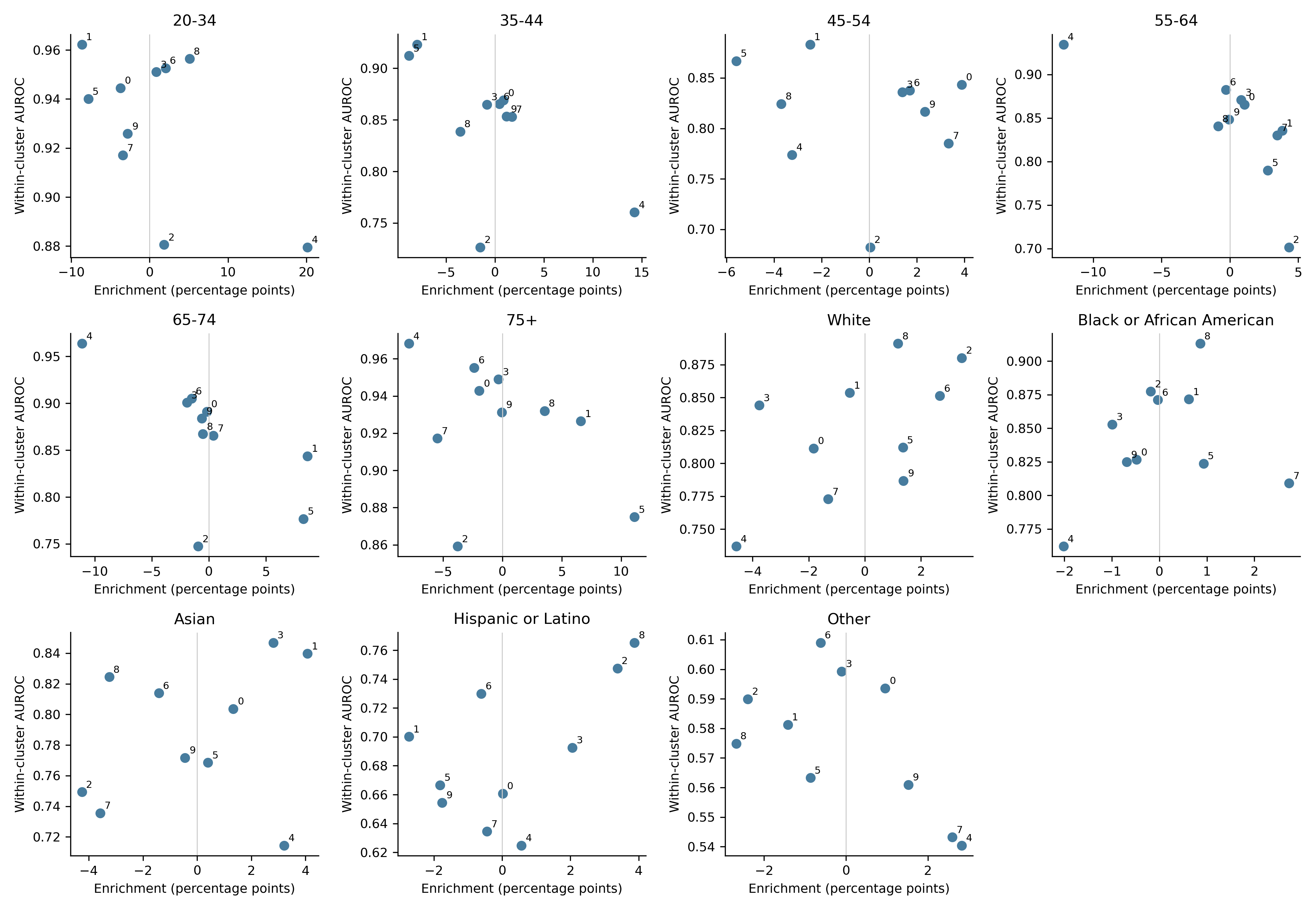}
  \caption{\textbf{Breast subgroup enrichment versus within-cluster AUROC.}
    Signed subgroup enrichment (percentage points relative to the eligible
    breast cohort) versus within-cluster subgroup AUROC under S1; each point
    is one canonical diagnosis cluster, labeled by cluster index. These are the per-cluster values
    underlying Fig.~\ref{fig:stratification}E.}
  \label{sfig:breast-enrichment-vs-auroc}
\end{figure}

\begin{figure}[htbp]
  \centering
  \includegraphics[width=\textwidth]{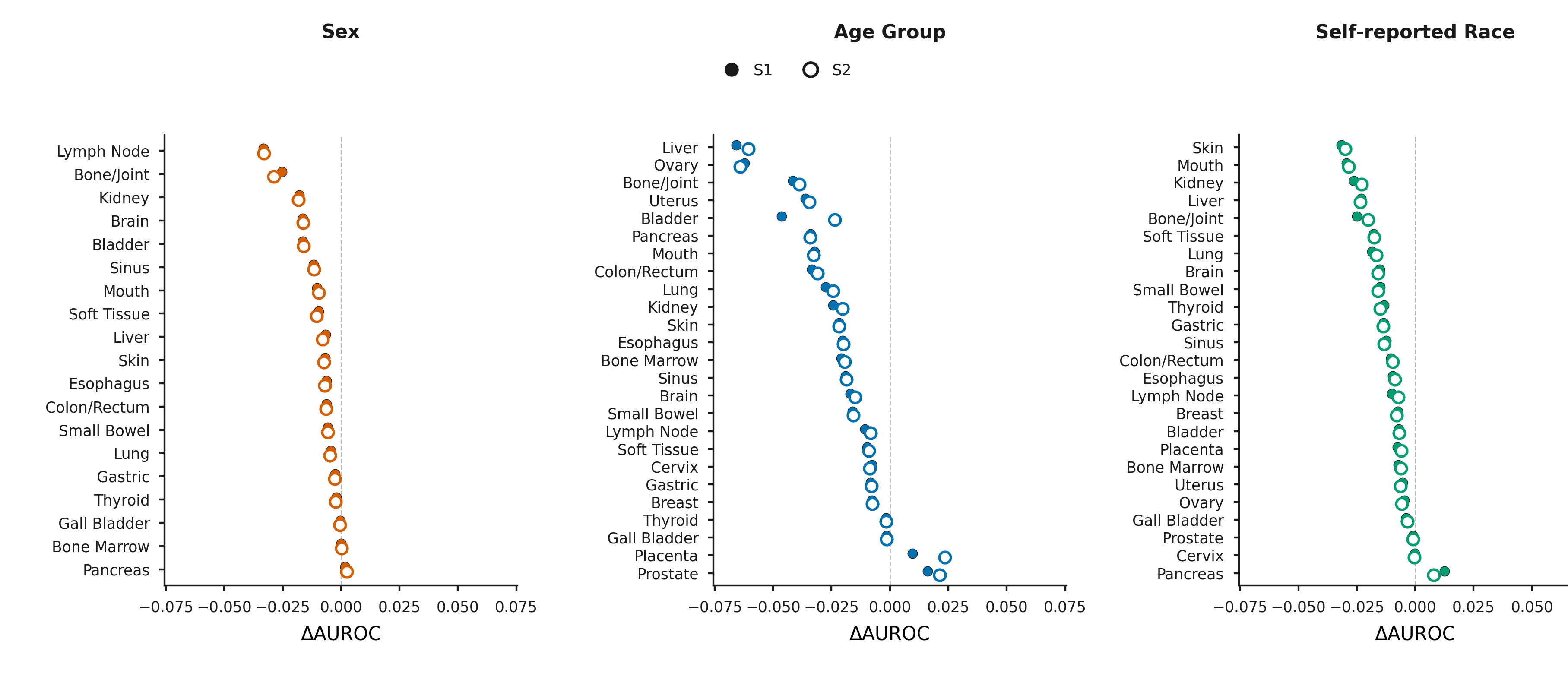}
  \caption{\textbf{Pan-organ cluster-stratified versus pooled AUROC.}
    Size-weighted pooled-versus-cluster-stratified AUROC under S1 across the
    69 organ--demographic combinations (25 age-group, 25 self-reported-race,
    and 19 sex). Per-organ values appear in Supplementary
    Tables~\ref{tab:supp-a4-aggregate}--\ref{tab:supp-a4-full}.}
  \label{sfig:panorgan-diagnosis-stratified}
\end{figure}

\clearpage

\subsection*{Cohort summary}

\begin{table}[H]
\centering
\caption{%
  \textbf{MSHS cohort summary.}
  Cohort hierarchy and demographic composition. Sex and self-reported race
  are patient-level; age is case-level. Sex and self-reported-race percentages use all
  257,322 patients, including unknown or excluded labels, as the denominator.
  Age-group percentages use the 318,174 cases with known age, excluding the
  ``Unknown/excluded'' row.
}
\label{tab:cohort}
\small
\setlength{\tabcolsep}{5pt}
\renewcommand{\arraystretch}{0.9}
\begin{tabular}{llrr}
\toprule
\textbf{Category} & \textbf{Subgroup} & \textbf{$n$} & \textbf{\%} \\
\midrule
\multicolumn{2}{l}{\textit{Cohort hierarchy}} & & \\
 & Patients & 257,322 & — \\
 & Cases & 327,483 & — \\
 & Specimens & 731,972 & — \\
 & Whole-slide images & 1,302,440 & — \\
\midrule
\multicolumn{2}{l}{\textit{Sex (patient level)}} & & \\
 & \textit{Total with valid sex} & \textit{255,382} & \\
 & Female          & 159,411 & 62.0 \\
 & Male            & 95,971  & 37.3 \\
 & Excluded (Indeterminate / Unknown) & 1,940 & 0.8 \\
\midrule
\multicolumn{2}{l}{\textit{Self-reported race (patient level)}} & & \\
 & \textit{Total with valid race} & \textit{207,500} & \\
 & White                       & 90,761 & 35.3 \\
 & Hispanic or Latino          & 47,220 & 18.4 \\
 & Black or African American   & 36,388 & 14.1 \\
 & Asian                       & 18,457 & 7.2  \\
 & Other$^{\dagger}$           & 14,674 & 5.7  \\
 & Excluded (Unknown)          & 49,822 & 19.4 \\
\midrule
\multicolumn{2}{l}{\textit{Age at case (continuous, case level)}} & & \\
 & Mean $\pm$ SD (years) & \multicolumn{2}{c}{$53.0 \pm 17.8$} \\
 & Median [IQR] & \multicolumn{2}{c}{54\ [40--67]} \\
\multicolumn{2}{l}{\textit{Age group (case level, model bins)$^{\ddagger}$}} & & \\
 & \textit{Total with known age} & \textit{318,174} & \\
 & 20--34 & 48,126 & 15.1 \\
 & 35--44 & 50,160 & 15.8 \\
 & 45--54 & 58,635 & 18.4 \\
 & 55--64 & 64,503 & 20.3 \\
 & 65--74 & 59,356 & 18.7 \\
 & $\geq$75 & 37,394 & 11.8 \\
 & Unknown/excluded & 9,309 & — \\
\midrule
\multicolumn{2}{l}{\textit{Cases per patient}} & & \\
 & Mean $\pm$ SD & $1.27 \pm 0.72$ & — \\
 & Median [IQR] & 1\ [1--1] & — \\
\multicolumn{2}{l}{\textit{Specimens per case}} & & \\
 & Mean $\pm$ SD & $2.24 \pm 2.17$ & — \\
 & Median [IQR] & 1\ [1--3] & — \\
\multicolumn{2}{l}{\textit{Slides per case}} & & \\
 & Mean $\pm$ SD & $4.27 \pm 7.27$ & — \\
 & Median [IQR] & 2\ [1--4] & — \\
\bottomrule
\multicolumn{4}{l}{\footnotesize $^{\dagger}$``Other'' consolidates Native Hawaiian or Other Pacific Islander,}\\
\multicolumn{4}{l}{\footnotesize \phantom{$^{\dagger}$}American Indian or Alaska Native, and Other Race.}\\
\multicolumn{4}{l}{\footnotesize $^{\ddagger}$Age groups include cases aged $\geq$20 years; ``Unknown/excluded''}\\
\multicolumn{4}{l}{\footnotesize \phantom{$^{\ddagger}$}includes cases with missing age or age $<$20 years.}\\
\end{tabular}
\end{table}

\clearpage
\subsection*{Per-organ cohort scale and composition}

\begin{table}[htbp]
  \centering\small\setlength{\tabcolsep}{7pt}
  \caption{\textbf{Per-organ cohort scale.}
    Unique patient, case, specimen, and slide counts for all 25 study organs
    in the MSHS 2025 cohort, ordered by descending slide count.
    Counts are derived after specimen organ normalization and slide-readiness
    filtering (feature and coordinate files present).}
  \label{tab:supp-cohort-scale}
  \begin{tabular}{lrrrr}
    \toprule
    \textbf{Organ} & \textbf{Patients} & \textbf{Cases} & \textbf{Specimens} & \textbf{Slides} \\
    \midrule
    Breast        & 37,426 &  46,145 &  75,781 & 263,230 \\
    Colon/Rectum  & 54,976 &  60,496 & 144,880 & 175,918 \\
    Prostate      &  8,814 &   9,880 &  58,135 & 142,270 \\
    Lymph Node    & 14,277 &  15,400 &  28,473 &  84,598 \\
    Uterus        & 17,014 &  19,795 &  22,915 &  79,630 \\
    Gastric       & 39,205 &  43,135 &  67,104 &  72,363 \\
    Skin          & 29,581 &  32,938 &  41,887 &  58,824 \\
    Placenta      & 13,766 &  14,592 &  14,795 &  50,946 \\
    Small Bowel   & 26,228 &  28,643 &  37,098 &  45,191 \\
    Thyroid       &  3,197 &   3,322 &   4,355 &  42,999 \\
    Cervix        & 10,831 &  13,337 &  30,815 &  35,644 \\
    Esophagus     & 18,726 &  20,689 &  31,458 &  31,559 \\
    Bone/Joint    & 12,344 &  13,808 &  16,869 &  30,352 \\
    Soft Tissue   & 10,099 &  10,684 &  13,684 &  28,628 \\
    Kidney        &  4,164 &   4,690 &   9,606 &  21,361 \\
    Bone Marrow   &  6,177 &  11,386 &  29,563 &  20,844 \\
    Mouth         &  6,759 &   7,431 &  12,953 &  19,877 \\
    Bladder       &  3,192 &   4,342 &   5,844 &  16,794 \\
    Ovary         &  2,553 &   2,592 &   3,487 &  14,689 \\
    Sinus         &  4,458 &   4,703 &  11,772 &  13,864 \\
    Lung          &  2,715 &   3,080 &   4,092 &  11,362 \\
    Liver         &  3,313 &   3,635 &   4,287 &  11,243 \\
    Gallbladder  &  5,506 &   5,517 &   5,562 &   9,928 \\
    Brain         &  2,323 &   2,458 &   4,144 &   8,926 \\
    Pancreas      &    814 &     925 &   1,009 &   6,089 \\
    \midrule
    \textit{Full cohort}
                  & \textit{257,322} & \textit{327,483} & \textit{731,972} & \textit{1,302,440} \\
    \bottomrule
  \end{tabular}
  \caption*{\footnotesize Full-cohort totals include records outside the
    25-organ, slide-ready population represented by the individual rows.
    Patient total reflects unique individuals in the full cohort;
    per-organ patient counts sum to more than the cohort total because patients
    contributing to multiple organ types are counted once per organ.}
\end{table}

\begin{table}[htbp]
  \centering\footnotesize\setlength{\tabcolsep}{4pt}
  \caption{\textbf{Per-organ patient-level demographic composition.}
    Within-organ percentages of patients by sex and self-reported race. The
    denominator is the full per-organ patient count (Supplementary
    Table~\ref{tab:supp-cohort-scale}); patients with other or unknown sex or
    unknown self-reported race are included in the denominator but not shown,
    so rows do not sum to 100\%, and the unknown fraction varies substantially
    across organs. Organs are ordered by descending slide count.}
  \label{tab:supp-demographic-composition}
  \begin{tabular}{lrrrrrrrr}
    \toprule
    & \multicolumn{2}{c}{\textbf{Sex (\%)}}
    & \multicolumn{5}{c}{\textbf{Self-reported Race (\%)}} \\
    \cmidrule(lr){2-3}\cmidrule(lr){4-8}
    \textbf{Organ} & \textbf{F} & \textbf{M}
    & \textbf{White} & \textbf{Hisp} & \textbf{Black} & \textbf{Asian} & \textbf{Other} \\
    \midrule
    Breast       & 94.7 &  1.0 & 25 & 12 & 11 &  6 & 4 \\
    Colon/Rectum & 48.1 & 51.7 & 38 & 18 & 14 &  6 & 6 \\
    Prostate     &  0.0 & 99.9 & 44 & 17 & 19 &  5 & 6 \\
    Lymph Node   & 60.2 & 38.1 & 38 & 18 & 17 & 12 & 4 \\
    Uterus       & 99.4 &  0.4 & 33 & 22 & 22 &  9 & 5 \\
    Gastric      & 58.1 & 41.8 & 31 & 25 & 16 &  6 & 6 \\
    Skin         & 51.1 & 48.5 & 40 & 14 & 11 &  5 & 5 \\
    Placenta     & 99.9 &  0.1 & 45 & 20 & 15 &  9 & 5 \\
    Small Bowel  & 55.3 & 44.6 & 38 & 18 & 13 &  5 & 6 \\
    Thyroid      & 75.0 & 24.9 & 37 & 20 & 13 & 21 & 4 \\
    Cervix       & 99.8 &  0.1 & 37 & 17 & 14 & 10 & 6 \\
    Esophagus    & 51.7 & 48.2 & 39 & 18 & 12 &  5 & 6 \\
    Bone/Joint   & 52.1 & 47.8 & 38 & 23 & 19 &  6 & 5 \\
    Soft Tissue  & 46.1 & 53.7 & 37 & 24 & 18 &  8 & 6 \\
    Kidney       & 41.1 & 58.8 & 32 & 24 & 23 & 10 & 6 \\
    Bone Marrow  & 45.9 & 54.0 & 41 & 18 & 19 &  7 & 8 \\
    Mouth        & 47.6 & 52.1 & 33 & 13 &  8 &  8 & 5 \\
    Bladder      & 26.3 & 73.6 & 51 & 18 & 14 &  7 & 5 \\
    Ovary        & 99.8 &  0.2 & 40 & 21 & 19 & 12 & 5 \\
    Sinus        & 39.7 & 60.1 & 44 & 17 &  9 & 10 & 7 \\
    Lung         & 50.2 & 49.6 & 34 & 22 & 24 &  9 & 7 \\
    Liver        & 53.0 & 46.8 & 32 & 26 & 17 & 10 & 7 \\
    Gallbladder & 68.5 & 31.4 & 27 & 38 & 12 &  9 & 8 \\
    Brain        & 54.2 & 45.7 & 32 & 22 & 19 & 11 & 6 \\
    Pancreas     & 48.2 & 51.8 & 37 & 25 & 20 & 10 & 4 \\
    \bottomrule
  \end{tabular}
  \caption*{\footnotesize F = Female; M = Male; Hisp = Hispanic or Latino;
    Black = Black or African American.}
\end{table}
\FloatBarrier

\subsection*{Diagnosis extraction}
\refstepcounter{suppnote}\label{suppnote:diagnosis-extraction}

\noindent\textbf{Supplementary Note~\ref{suppnote:diagnosis-extraction}.}
Specimen-level diagnoses were extracted from free-text pathology reports with a
large language model (Azure OpenAI, GPT-5 mini) returning output constrained to
a JSON schema. For each report the model emitted a list of specimens; each
specimen carried an identifier (the specimen letter, e.g.\ ``A''), an anatomic
source, a normalized organ label drawn from the 25 study organ groups plus an
\emph{Other} category, and a diagnosis string. The instruction protocol
directed the model to: (i)~split each report into its constituent specimens
using the explicit specimen headers; (ii)~take the anatomic source and
diagnosis preferentially from the \emph{Diagnosis} / \emph{Final Diagnosis}
section, falling back to the specimen-source field only when the location was
absent there; (iii)~keep the diagnosis text \emph{verbatim}, concatenating
relevant \emph{Addendum}, \emph{Comment}, and \emph{Note} sections while
deduplicating repeated statements; (iv)~resolve conflicts by the priority
Final Diagnosis $>$ Addendum $>$ Comment $>$ Note; and (v)~never invent fields
or facts, ignoring billing codes and boilerplate. Only the extracted
specimen-level diagnosis text was used downstream; no protected health
information was retained in the derived data.

\subsection*{Diagnostic spectrum embedding}
\refstepcounter{suppnote}\label{suppnote:dx-spectrum}

\noindent\textbf{Supplementary Note~\ref{suppnote:dx-spectrum}.}
The diagnostic-spectrum map in Fig.~\ref{fig:framework}B was generated from
the extracted specimen-level diagnosis text
(Supplementary Note~\ref{suppnote:diagnosis-extraction}). Of 731,204 specimens
with non-empty diagnosis text, 51,264 mapped to the \emph{Other} organ category
or lacked an organ label and were excluded, leaving 679,940 specimens from the
25 study organs. Diagnosis strings were embedded with Azure OpenAI
\texttt{text-embedding-3-large} (3,072 dimensions). Each specimen was assigned
to one of four broad categories by a rule-based mapping of the diagnosis text:
neoplastic ($n=176{,}625$; 26.0\%), normal ($n=166{,}658$; 24.5\%),
inflammatory, including infectious findings ($n=143{,}816$; 21.2\%), and other
($n=192{,}841$; 28.4\%), comprising non-neoplastic structural findings, tissue
injury or degenerative changes, and diagnoses not matched by any rule (56.6\%
of the other category). To prevent high-volume organs and categories from
dominating the layout, a reference set of 99,949 specimens was sampled with
balanced quotas across organ $\times$ category cells (up to 1,225 per cell).
Embeddings were reduced to 50 principal components (65.7\% of variance
explained), and a single UMAP was fitted on the reference set (cosine
distance, 50 neighbours, minimum distance 0.5, fixed random seed); the
remaining 579,991 specimens were projected into the fitted embedding. Because
templated sign-out text yields near-identical embeddings, dense islands
reflect reporting conventions rather than prevalence. The 20 labelled example
diagnoses (five per category) were reviewed by pathologists, and the category
of the displayed labels follows that review; the rule-based assignment used for
the percentages was not modified, so a small number of individual specimens
may be categorized differently from the pathologists' assessment.

\section*{Supplementary Methods}
\addcontentsline{toc}{section}{Supplementary Methods}

\suppmethodheading{Grouped training and epoch exposure}{supp:training-implementation}

Grouped balancing for S3 and S3$_\text{RND}$ operated on unique pathology cases,
with each case counted once in the candidate training pool. Within a grouping
pool $p$, let $n_{pg}$ denote the
number of cases in demographic class $g$. The balancing target was the largest
available class count,
$n_p^{\max}=\max_{g\in\{1,\ldots,G\}} n_{pg}$.
A class was retained only when $n_p^{\max}/n_{pg}\leq 50$; classes requiring
greater oversampling were dropped, and a pool with fewer than two retained classes was excluded from
group construction. Classes below $n_p^{\max}$ were oversampled by cycling
through shuffled case lists to that target; the largest class was retained in
full, without subsampling majority-class cases. The
balanced class lists were assembled into training groups containing one case
from each of the $G$ retained demographic classes. Each case was then
represented by its full case-level MIL bag.

S3 and its random grouped-training control used this formulation. S3$_\text{RND}$ applied
grouped demographic balancing once to the full, undifferentiated training pool;
it did not use diagnosis clusters or randomized cluster assignments and served
as the non-semantic global grouped-balancing control. S3 applied the
same formulation separately within the diagnosis-context clusters defined
above. Clusters with fewer than two retained demographic classes were
excluded, and groups from the retained clusters were combined for training. S3$_\text{RND}$ is reported as a supporting control to S3.

S1 and S2 processed cases individually and computed loss case by case. Each S3 or S3$_\text{RND}$
training example was a demographic group; the case-level cross-entropy losses
were summed across group members before gradient accumulation and the optimizer
step. Thus, although all regimes retained the same case-level MIL prediction
unit, grouped training changed the effective batch structure and case presentations
per epoch relative to S2. Grouped training used a gradient-accumulation setting of 8.

The S1 per-epoch case target was
$N_\text{epoch}=\min\{N_\text{train},\max[\operatorname{round}(0.5N_\text{train}),11{,}018]\}$. S2 sampled $\max\{32,\operatorname{round}(N_\text{epoch}/G)\}$
cases per target class, where $G$ is the number of target classes. Both grouped regimes instead
capped the number of groups presented per epoch using
a tiered rule: all groups when at most 512 were available; 256 groups
when at most 2,048 were available; 512 groups when at most 8,192 were available;
and 1,024 groups for larger pools. These dataset-size-aware caps were
computational design choices, not scientific controls, and contribute to the
exposure difference between S2 and S3.

\paragraph{Training-side cluster-assignment sensitivity.}
Ties in case-level cluster voting were resolved by specimen row order,
affecting $0.3$--$35.2\%$ of training cases across organ--demographic
combinations. Rebuilding groups with alternative deterministic tie-breaking
rules retained at least $99.6\%$ training-case overlap. Relabeling production
groups using the lowest tied cluster index increased the mean residual
class--cluster Cram\'{e}r's $V$ by $0.016$, compared with $0.146$ in the
unbalanced pool. This group-construction check did not retrain models or
establish an effect on AUROC.

\suppmethodheading{Cluster-stratified evaluation}{supp:cluster-evaluation}

For cluster-stratified evaluation, specimen-level assignments were aggregated
by majority vote; ties were resolved by the highest summed centroid similarity,
with any remaining tie resolved by the lowest cluster index.
The canonical breast evaluation retained 5,112 tied cases. This differs from
S3 training assignment, which resolves ties by specimen row order.

As a post-hoc check on diagnostic composition, AUROC was recomputed \emph{within}
diagnosis clusters instead of over the pooled validation set. Let $A_k$ denote the
macro one-vs.-rest AUROC computed within cluster $k$, restricted to the classes
present in that cluster; a cluster was excluded only when a single ground-truth
class was present, which leaves AUROC undefined. Writing $n_k$ for the number of
validation cases in cluster $k$, we compare
\begin{equation*}
\text{AUROC}_\text{pooled}
\qquad\text{against}\qquad
\text{AUROC}_\text{strat} = \sum_k \frac{n_k}{\sum_j n_j}\, A_k ,
\end{equation*}
where $\text{AUROC}_\text{pooled}$ is the ordinary macro AUROC over all validation
cases and $\text{AUROC}_\text{strat}$ averages the within-cluster values in
proportion to cluster size. The pooled statistic includes rankings between
cases from different clusters, whereas $\text{AUROC}_\text{strat}$ contains
only within-cluster rankings. The primary cluster-stratified contrast was
\begin{equation*}
\Delta = \text{AUROC}_\text{strat} - \text{AUROC}_\text{pooled},
\end{equation*}
tested per organ with a paired $t$-test across the 10 matched MCCV folds. The
same calculation after random reassignment of cluster labels served as a
separate evaluation-only random-label control; it is unrelated to the
S3$_\text{RND}$ training regime. Drop\,\% (defined as a fraction of above-chance pooled performance) is
reported per organ in Supplementary Table~\ref{tab:supp-a4-full}. $\Delta$
remained the primary contrast measure throughout. Case-to-cluster matching was
at least 99.4\% complete for every organ and fold (lowest, pancreas), and
pooled and stratified AUROC were computed over the same matched cases.

\suppdetailheading{Breast worked example: pooled versus cluster-stratified AUROC}{supp:breast-evaluation}

Fig.~\ref{fig:stratification}D reports $\text{AUROC}_\text{pooled}$,
$\text{AUROC}_\text{strat}$ and $\Delta$ exactly as defined above, taken from
the same per-fold pan-organ evaluation that underlies Supplementary
Tables~\ref{tab:supp-a4-aggregate}--\ref{tab:supp-a4-full} (selected $K=10$);
the breast values therefore coincide with the breast row of Supplementary
Table~\ref{tab:supp-a4-full}. In every fold, all ten breast clusters contained
at least two target classes, so all matched validation cases ($8{,}747$ for
age group; $5{,}760$--$5{,}901$ for self-reported race) contributed to both
statistics.

For Fig.~\ref{fig:stratification}B, UMAP visualizes fixed $k$-means
assignments; it does not define the clusters. In panel E, signed enrichment is
the subgroup's within-cluster proportion minus its overall breast proportion,
in percentage points. Spearman correlations use the corresponding
within-cluster one-vs.-rest AUROC under S1. Enrichment uses all cases with
known demographic labels, whereas AUROC uses held-out cases with the same
canonical assignments. Leave-one-cluster-out ranges describe
sensitivity to individual clusters, not confidence intervals.

\suppmethodheading{LLM-assisted literature searching and hypothesis generation}{supp:hypothesis-map}

\paragraph{Model, inputs, and search procedure.}
We used OpenAI \texttt{gpt-6-astra} on 24 September 2026 (UTC) to examine
prediction of self-reported race in skin, age-group prediction in breast,
and sex prediction in gastric tissue. A fixed prompt was applied separately
to each example, with the organ, demographic target, and observed model
results supplied as inputs. For skin, the supplied results comprised residual
S3 AUROC, best-context AUROC, and their difference; other
strategy-level values were verified subsequently and were not generation
inputs. The three candidates were selected based on residual predictability,
clinical relevance, and potential for histomorphological investigation.
Context-audit values were supplied as supplementary information during
literature hypothesis generation; image--context gaps were not formal
candidate-selection criteria. The prompt below is a reformatted, reproducible
version of the search instructions.

Retrieval used web search and NCBI PubMed ESearch/EFetch, with PMC EFetch where
full text was accessible. The generation stage proposed candidate tissue or
contextual explanations and linked supporting claims to publications.
Contradictory and null evidence was sought alongside supporting evidence.
The search was targeted and exploratory, not exhaustive or systematic.

\paragraph{Source checking and interpretation.}
In a separate pass, the same model checked citation identity with PubMed
ESummary and assessed each claim against the retrieved abstract or accessible
publisher or PMC full text. It recorded the source identifier, material
reviewed, supporting passage where available, and limits on applicability.
Claims were narrowed or excluded when the inspected material did not support
them. DOI and PMID records were checked against Crossref and PubMed. Claim
support was assessed by the model without independent author verification.

\paragraph{Applicability of the evidence.}
Skin pigmentation studies differed in imaging modality, population, anatomical
site, and exposure, limiting transfer to routine H\&E in this cohort. Breast
morphometry provided cross-sectional evidence on tissue composition rather
than longitudinal evidence of adipose replacement\cite{sandhu2016morphometry}.
Gastric findings were specific to disease subtype, subsite, and treatment.
Signet-ring and diffuse categories were distinguished, as were within-cancer
subtype proportions and population incidence. The review retained null findings,
including no significant sex difference in atrophy scores in the eradication
study\cite{kodama2021mucosal}, similar sex-specific rates of diffuse gastric
carcinoma\cite{janssen1991gastric,derakhshan2009oesophageal}, no adjusted
association between quantitative involution measures and subsequent breast
cancer risk\cite{kensler2020tdlu}, and no observed difference in dermal
collagen or elastic-fibre organisation between the skin types compared\cite{girardeau2009dermal}.

\paragraph{Reproducible search prompt.}
\begin{quote}
\small
\textbf{Study question.} Organ: [organ]. Demographic target: [target].
Observed model results: [supplied results]. What clinical, disease-context,
workflow, or histomorphologic factors could plausibly contribute to the
observed demographic-associated signal in routine H\&E histopathology?
The observed signal does not establish its source. Generate testable
hypotheses, not a conclusive explanation.

\textbf{Search scope.} Search clinical and contextual explanations, including
disease prevalence or subtype, diagnosis distribution, anatomic subsite,
procedure or sampling, severity, treatment history, clinical workflow,
exposure, and healthcare context. Also search histomorphologic explanations,
including H\&E-visible epithelial or cellular morphology, stroma, immune
composition, fibrosis, pigmentation, adipose composition, vascular features,
age-related remodeling, and other tissue features. Specify whether each
proposed feature is localized to a compartment, lesion, or structure; diffuse
across broader tissue architecture; or mixed. Include technical or workflow
explanations when directly relevant, without requiring a separate category.
Do not treat self-reported race as a direct biological variable. Consider
environment, exposure, healthcare access, geography, disease distribution,
ancestry-correlated factors, institutional context, and workflow where
supported.

\textbf{Evidence requirements.} Prefer peer-reviewed primary human pathology
or histology studies, other direct human tissue studies, and strong pathology
reviews; use epidemiologic or clinical studies for prevalence or context
hypotheses. Label findings from molecular, imaging, animal, or mechanistic
studies as indirect supporting context. Look for relevant contradictory and
null findings. Do not generalize across organs, populations, demographic
definitions, or disease settings without labeling the extrapolation as
indirect. For every retained claim, identify its source, provide a DOI, PMID,
or stable URL where available, identify the supporting location where
available (abstract, section, page, figure, or table), and paraphrase what the
source supports. Distinguish source support from the hypothesis. Do not
invent citations or supporting locations. State when only the abstract is
accessible and restrict claims accordingly. If no useful morphology
hypothesis is supported, say so. Do not assign arbitrary evidence scores or
high, medium, or low confidence labels.

\textbf{Requested output.} Briefly restate the supplied observed signal
without adding findings. Aim for two to five useful candidate hypotheses,
returning fewer or none if warranted. For each, report its type (clinical or
contextual, morphology, or workflow/technical), hypothesis, spatial scope
(localized, diffuse, mixed, or not applicable), features to inspect, plausible
relation to the observed signal, supporting claims and sources, alternative
explanations, and remaining evidence gaps. For each supporting source,
provide its citation and available identifiers, evidence type and indirectness,
actual access status (open full text, institutional full text, abstract only,
or not reviewed), available supporting locations, and what they support.
Link each claim to its source and location. List each paper once, allowing
multiple claim links. Use short paraphrases rather than long quotations.
Finish with unresolved questions suitable for matching, stratification,
measurement, or tissue-review checks.

\textbf{Style and limits.} Use concise scientific language understandable to
a pathology or biomedical research collaborator. Prefer a few traceable
hypotheses over broad coverage. Do not claim causality, call a candidate
feature a mechanism unless directly demonstrated, treat model reasoning as
evidence, attempt an exhaustive systematic review, or force a hypothesis
when evidence is weak.
\end{quote}

\suppmethodheading{Age-class harmonization}{supp:design-sensitivity}

\suppdetailheading{Class-set and validation comparability}{supp:class-comparability}

For prostate and placenta age prediction, rare age groups were excluded
under the S3 class-retention rule. Prostate S3$_\text{RND}$ and S3
originally used different class sets; their comparison was recomputed using
four common age classes (45--54, 55--64, 65--74, and 75+) and 1,954 matched
validation cases in each of the ten MCCV folds. Placenta used three common
classes (20--34, 35--44, and 45--54) and 2,872 matched cases per fold; its S3
comparison was unchanged. All models were evaluated at epoch 50. Macro
one-vs.-rest AUROC was averaged over the common classes using their original
softmax probabilities, without renormalization. Other organ-level comparisons
retained their original evaluation definitions.

For prostate age, harmonized AUROC was $0.714\pm0.008$ under
S3$_\text{RND}$ and $0.700\pm0.010$ under S3 (mean $\pm$ SD
across ten folds). The paired increment was $0.0140\pm0.0053$, positive in
all ten folds (paired $t$-test $p=1.5\times10^{-5}$).

Figure~\ref{fig:s3} and Supplementary Table~\ref{tab:supp-s3-increment}
use the harmonized prostate S3 comparison. Supplementary
Table~\ref{tab:supp-full-ladder} retains original per-regime AUROCs.
S1/S2-to-S3 comparisons for prostate and placenta remain non-class-matched;
no harmonized cross-stage contrasts are reported here.

\section*{Supplementary results}

\begin{table}[htbp]
\centering
\caption{\textbf{Per-organ macro one-vs.-rest AUROC across training regimes.}
  Mean${\scriptstyle\pm}$SD across 10 MCCV folds under S1, S2,
  S3\textsubscript{RND}, and S3
  (\headingref{sec:methods-training-regimes}). Dashes (---) mark sex-specific
  organs excluded from the sex target; bold entries follow the top-performer
  rule of Fig.~\ref{fig:results}.}
\label{tab:supp-full-ladder}
\scriptsize
\setlength{\tabcolsep}{2.2pt}
\renewcommand{\arraystretch}{0.95}
\resizebox{\textwidth}{!}{%
\begin{tabular}{lcccccccccccc}
\toprule
 & \multicolumn{4}{c}{\textbf{Sex}} & \multicolumn{4}{c}{\textbf{Age Group}} & \multicolumn{4}{c}{\textbf{Self-reported Race}} \\
\cmidrule(lr){2-5}\cmidrule(lr){6-9}\cmidrule(lr){10-13}
\textbf{Organ} & S1 & S2 & S3$_\text{RND}$ & S3 & S1 & S2 & S3$_\text{RND}$ & S3 & S1 & S2 & S3$_\text{RND}$ & S3 \\
\midrule
Bladder      & $0.762{\scriptstyle\pm0.026}$ & $0.761{\scriptstyle\pm0.029}$ & $0.731{\scriptstyle\pm0.017}$ & $0.724{\scriptstyle\pm0.014}$ & $0.656{\scriptstyle\pm0.012}$ & $0.614{\scriptstyle\pm0.018}$ & $0.595{\scriptstyle\pm0.013}$ & $0.583{\scriptstyle\pm0.019}$ & $0.566{\scriptstyle\pm0.012}$ & $0.548{\scriptstyle\pm0.018}$ & $0.547{\scriptstyle\pm0.012}$ & $0.541{\scriptstyle\pm0.011}$ \\
Bone Marrow  & $0.773{\scriptstyle\pm0.022}$ & $0.767{\scriptstyle\pm0.019}$ & $0.607{\scriptstyle\pm0.014}$ & $0.607{\scriptstyle\pm0.013}$ & $0.687{\scriptstyle\pm0.008}$ & $0.666{\scriptstyle\pm0.012}$ & $0.621{\scriptstyle\pm0.008}$ & $0.611{\scriptstyle\pm0.008}$ & $0.645{\scriptstyle\pm0.014}$ & $0.628{\scriptstyle\pm0.014}$ & $0.592{\scriptstyle\pm0.008}$ & $0.585{\scriptstyle\pm0.008}$ \\
Bone/Joint   & $0.759{\scriptstyle\pm0.017}$ & $0.746{\scriptstyle\pm0.014}$ & $0.650{\scriptstyle\pm0.008}$ & $0.615{\scriptstyle\pm0.008}$ & $0.709{\scriptstyle\pm0.006}$ & $0.700{\scriptstyle\pm0.007}$ & $0.660{\scriptstyle\pm0.006}$ & $0.615{\scriptstyle\pm0.007}$ & $0.637{\scriptstyle\pm0.011}$ & $0.606{\scriptstyle\pm0.014}$ & $0.577{\scriptstyle\pm0.009}$ & $0.546{\scriptstyle\pm0.010}$ \\
Brain        & $0.604{\scriptstyle\pm0.021}$ & $0.603{\scriptstyle\pm0.023}$ & $0.534{\scriptstyle\pm0.022}$ & $0.514{\scriptstyle\pm0.024}$ & $0.601{\scriptstyle\pm0.014}$ & $0.599{\scriptstyle\pm0.015}$ & $0.590{\scriptstyle\pm0.020}$ & $0.574{\scriptstyle\pm0.020}$ & $0.540{\scriptstyle\pm0.013}$ & $0.536{\scriptstyle\pm0.014}$ & $0.542{\scriptstyle\pm0.016}$ & $0.523{\scriptstyle\pm0.015}$ \\
Breast       & --- & --- & --- & --- & $\mathbf{0.882}{\scriptstyle\pm0.002}$ & $\mathbf{0.878}{\scriptstyle\pm0.002}$ & $\mathbf{0.818}{\scriptstyle\pm0.002}$ & $\mathbf{0.816}{\scriptstyle\pm0.002}$ & $0.750{\scriptstyle\pm0.007}$ & $0.726{\scriptstyle\pm0.006}$ & $0.646{\scriptstyle\pm0.005}$ & $0.642{\scriptstyle\pm0.005}$ \\
Cervix       & --- & --- & --- & --- & $0.834{\scriptstyle\pm0.005}$ & $0.815{\scriptstyle\pm0.006}$ & $0.788{\scriptstyle\pm0.004}$ & $\mathbf{0.783}{\scriptstyle\pm0.004}$ & $0.647{\scriptstyle\pm0.008}$ & $0.633{\scriptstyle\pm0.008}$ & $0.607{\scriptstyle\pm0.008}$ & $0.605{\scriptstyle\pm0.007}$ \\
Colon/Rectum & $\mathbf{0.965}{\scriptstyle\pm0.002}$ & $\mathbf{0.962}{\scriptstyle\pm0.002}$ & $0.727{\scriptstyle\pm0.004}$ & $0.728{\scriptstyle\pm0.003}$ & $0.784{\scriptstyle\pm0.004}$ & $0.775{\scriptstyle\pm0.007}$ & $0.701{\scriptstyle\pm0.002}$ & $0.670{\scriptstyle\pm0.002}$ & $\mathbf{0.759}{\scriptstyle\pm0.007}$ & $\mathbf{0.737}{\scriptstyle\pm0.008}$ & $0.655{\scriptstyle\pm0.002}$ & $0.645{\scriptstyle\pm0.004}$ \\
Esophagus    & $\mathbf{0.872}{\scriptstyle\pm0.008}$ & $\mathbf{0.852}{\scriptstyle\pm0.010}$ & $0.725{\scriptstyle\pm0.005}$ & $0.726{\scriptstyle\pm0.006}$ & $0.692{\scriptstyle\pm0.006}$ & $0.695{\scriptstyle\pm0.007}$ & $0.648{\scriptstyle\pm0.005}$ & $0.633{\scriptstyle\pm0.005}$ & $0.682{\scriptstyle\pm0.010}$ & $0.659{\scriptstyle\pm0.008}$ & $0.620{\scriptstyle\pm0.006}$ & $0.613{\scriptstyle\pm0.007}$ \\
Gallbladder & $\mathbf{0.949}{\scriptstyle\pm0.010}$ & $\mathbf{0.951}{\scriptstyle\pm0.009}$ & $0.805{\scriptstyle\pm0.019}$ & $0.804{\scriptstyle\pm0.021}$ & $0.790{\scriptstyle\pm0.022}$ & $0.790{\scriptstyle\pm0.023}$ & $0.667{\scriptstyle\pm0.007}$ & $0.668{\scriptstyle\pm0.006}$ & $0.669{\scriptstyle\pm0.012}$ & $0.660{\scriptstyle\pm0.010}$ & $0.596{\scriptstyle\pm0.009}$ & $0.593{\scriptstyle\pm0.009}$ \\
Gastric      & $\mathbf{0.970}{\scriptstyle\pm0.003}$ & $\mathbf{0.961}{\scriptstyle\pm0.005}$ & $\mathbf{0.861}{\scriptstyle\pm0.004}$ & $\mathbf{0.865}{\scriptstyle\pm0.004}$ & $0.804{\scriptstyle\pm0.007}$ & $0.804{\scriptstyle\pm0.007}$ & $0.696{\scriptstyle\pm0.002}$ & $0.688{\scriptstyle\pm0.003}$ & $\mathbf{0.776}{\scriptstyle\pm0.009}$ & $\mathbf{0.765}{\scriptstyle\pm0.010}$ & $\mathbf{0.700}{\scriptstyle\pm0.004}$ & $\mathbf{0.693}{\scriptstyle\pm0.005}$ \\
Kidney       & $0.777{\scriptstyle\pm0.015}$ & $0.776{\scriptstyle\pm0.016}$ & $0.681{\scriptstyle\pm0.013}$ & $0.680{\scriptstyle\pm0.010}$ & $0.682{\scriptstyle\pm0.010}$ & $0.683{\scriptstyle\pm0.010}$ & $0.618{\scriptstyle\pm0.012}$ & $0.589{\scriptstyle\pm0.010}$ & $0.602{\scriptstyle\pm0.016}$ & $0.589{\scriptstyle\pm0.019}$ & $0.579{\scriptstyle\pm0.010}$ & $0.545{\scriptstyle\pm0.010}$ \\
Liver        & $0.787{\scriptstyle\pm0.016}$ & $0.788{\scriptstyle\pm0.018}$ & $0.644{\scriptstyle\pm0.023}$ & $0.643{\scriptstyle\pm0.020}$ & $0.669{\scriptstyle\pm0.008}$ & $0.668{\scriptstyle\pm0.008}$ & $0.646{\scriptstyle\pm0.011}$ & $0.602{\scriptstyle\pm0.009}$ & $0.605{\scriptstyle\pm0.012}$ & $0.599{\scriptstyle\pm0.012}$ & $0.578{\scriptstyle\pm0.014}$ & $0.553{\scriptstyle\pm0.012}$ \\
Lung         & $0.660{\scriptstyle\pm0.016}$ & $0.662{\scriptstyle\pm0.016}$ & $0.599{\scriptstyle\pm0.017}$ & $0.594{\scriptstyle\pm0.017}$ & $0.732{\scriptstyle\pm0.009}$ & $0.716{\scriptstyle\pm0.009}$ & $0.661{\scriptstyle\pm0.011}$ & $0.653{\scriptstyle\pm0.014}$ & $0.626{\scriptstyle\pm0.014}$ & $0.613{\scriptstyle\pm0.012}$ & $0.598{\scriptstyle\pm0.012}$ & $0.582{\scriptstyle\pm0.009}$ \\
Lymph Node   & $\mathbf{0.960}{\scriptstyle\pm0.007}$ & $\mathbf{0.961}{\scriptstyle\pm0.006}$ & $\mathbf{0.854}{\scriptstyle\pm0.006}$ & $0.827{\scriptstyle\pm0.007}$ & $0.789{\scriptstyle\pm0.011}$ & $0.786{\scriptstyle\pm0.009}$ & $0.669{\scriptstyle\pm0.006}$ & $0.659{\scriptstyle\pm0.006}$ & $0.715{\scriptstyle\pm0.009}$ & $0.685{\scriptstyle\pm0.012}$ & $0.630{\scriptstyle\pm0.006}$ & $0.624{\scriptstyle\pm0.006}$ \\
Mouth        & $0.812{\scriptstyle\pm0.016}$ & $0.791{\scriptstyle\pm0.018}$ & $0.612{\scriptstyle\pm0.016}$ & $0.609{\scriptstyle\pm0.015}$ & $0.691{\scriptstyle\pm0.012}$ & $0.691{\scriptstyle\pm0.011}$ & $0.652{\scriptstyle\pm0.006}$ & $0.619{\scriptstyle\pm0.005}$ & $0.612{\scriptstyle\pm0.011}$ & $0.592{\scriptstyle\pm0.009}$ & $0.588{\scriptstyle\pm0.014}$ & $0.546{\scriptstyle\pm0.012}$ \\
Ovary        & --- & --- & --- & --- & $\mathbf{0.850}{\scriptstyle\pm0.007}$ & $\mathbf{0.845}{\scriptstyle\pm0.008}$ & $\mathbf{0.805}{\scriptstyle\pm0.010}$ & $0.771{\scriptstyle\pm0.016}$ & $0.588{\scriptstyle\pm0.013}$ & $0.577{\scriptstyle\pm0.022}$ & $0.555{\scriptstyle\pm0.018}$ & $0.550{\scriptstyle\pm0.015}$ \\
Pancreas     & $0.594{\scriptstyle\pm0.036}$ & $0.596{\scriptstyle\pm0.038}$ & $0.567{\scriptstyle\pm0.022}$ & $0.568{\scriptstyle\pm0.023}$ & $0.545{\scriptstyle\pm0.037}$ & $0.549{\scriptstyle\pm0.038}$ & $0.527{\scriptstyle\pm0.037}$ & $0.507{\scriptstyle\pm0.030}$ & $0.526{\scriptstyle\pm0.041}$ & $0.518{\scriptstyle\pm0.029}$ & $0.521{\scriptstyle\pm0.023}$ & $0.521{\scriptstyle\pm0.019}$ \\
Placenta     & --- & --- & --- & --- & $0.615{\scriptstyle\pm0.083}$ & $0.577{\scriptstyle\pm0.070}$ & $0.574{\scriptstyle\pm0.017}$ & $0.572{\scriptstyle\pm0.017}$ & $0.734{\scriptstyle\pm0.012}$ & $0.705{\scriptstyle\pm0.012}$ & $0.622{\scriptstyle\pm0.008}$ & $0.621{\scriptstyle\pm0.010}$ \\
Prostate     & --- & --- & --- & --- & $0.765{\scriptstyle\pm0.063}$ & $0.751{\scriptstyle\pm0.083}$ & $0.714{\scriptstyle\pm0.008}$ & $0.650{\scriptstyle\pm0.022}$ & $0.740{\scriptstyle\pm0.015}$ & $0.707{\scriptstyle\pm0.012}$ & $0.627{\scriptstyle\pm0.009}$ & $0.622{\scriptstyle\pm0.007}$ \\
Sinus        & $\mathbf{0.879}{\scriptstyle\pm0.013}$ & $\mathbf{0.880}{\scriptstyle\pm0.012}$ & $0.764{\scriptstyle\pm0.017}$ & $0.764{\scriptstyle\pm0.017}$ & $0.759{\scriptstyle\pm0.010}$ & $0.749{\scriptstyle\pm0.013}$ & $0.660{\scriptstyle\pm0.008}$ & $0.645{\scriptstyle\pm0.008}$ & $0.632{\scriptstyle\pm0.010}$ & $0.616{\scriptstyle\pm0.010}$ & $0.567{\scriptstyle\pm0.015}$ & $0.553{\scriptstyle\pm0.015}$ \\
Skin         & $\mathbf{0.912}{\scriptstyle\pm0.006}$ & $\mathbf{0.899}{\scriptstyle\pm0.007}$ & $0.763{\scriptstyle\pm0.004}$ & $0.765{\scriptstyle\pm0.004}$ & $0.766{\scriptstyle\pm0.003}$ & $0.766{\scriptstyle\pm0.002}$ & $0.674{\scriptstyle\pm0.003}$ & $0.658{\scriptstyle\pm0.003}$ & $\mathbf{0.793}{\scriptstyle\pm0.006}$ & $\mathbf{0.767}{\scriptstyle\pm0.006}$ & $\mathbf{0.758}{\scriptstyle\pm0.007}$ & $\mathbf{0.728}{\scriptstyle\pm0.006}$ \\
Small Bowel  & $\mathbf{0.942}{\scriptstyle\pm0.007}$ & $\mathbf{0.925}{\scriptstyle\pm0.009}$ & $0.785{\scriptstyle\pm0.003}$ & $0.791{\scriptstyle\pm0.004}$ & $0.749{\scriptstyle\pm0.006}$ & $0.749{\scriptstyle\pm0.006}$ & $0.662{\scriptstyle\pm0.003}$ & $0.647{\scriptstyle\pm0.004}$ & $0.726{\scriptstyle\pm0.008}$ & $0.709{\scriptstyle\pm0.011}$ & $0.665{\scriptstyle\pm0.005}$ & $\mathbf{0.647}{\scriptstyle\pm0.004}$ \\
Soft Tissue  & $0.823{\scriptstyle\pm0.010}$ & $0.805{\scriptstyle\pm0.011}$ & $0.658{\scriptstyle\pm0.009}$ & $0.649{\scriptstyle\pm0.009}$ & $0.659{\scriptstyle\pm0.009}$ & $0.663{\scriptstyle\pm0.008}$ & $0.601{\scriptstyle\pm0.006}$ & $0.585{\scriptstyle\pm0.006}$ & $0.605{\scriptstyle\pm0.010}$ & $0.583{\scriptstyle\pm0.006}$ & $0.566{\scriptstyle\pm0.007}$ & $0.555{\scriptstyle\pm0.005}$ \\
Thyroid      & $\mathbf{0.931}{\scriptstyle\pm0.016}$ & $\mathbf{0.936}{\scriptstyle\pm0.016}$ & $0.761{\scriptstyle\pm0.019}$ & $0.762{\scriptstyle\pm0.020}$ & $0.770{\scriptstyle\pm0.012}$ & $0.774{\scriptstyle\pm0.011}$ & $0.701{\scriptstyle\pm0.008}$ & $0.702{\scriptstyle\pm0.009}$ & $0.714{\scriptstyle\pm0.014}$ & $0.700{\scriptstyle\pm0.017}$ & $0.635{\scriptstyle\pm0.015}$ & $0.625{\scriptstyle\pm0.019}$ \\
Uterus       & --- & --- & --- & --- & $\mathbf{0.850}{\scriptstyle\pm0.003}$ & $\mathbf{0.842}{\scriptstyle\pm0.004}$ & $0.801{\scriptstyle\pm0.002}$ & $\mathbf{0.786}{\scriptstyle\pm0.003}$ & $0.653{\scriptstyle\pm0.005}$ & $0.644{\scriptstyle\pm0.005}$ & $0.608{\scriptstyle\pm0.005}$ & $0.604{\scriptstyle\pm0.004}$ \\
\midrule
\textbf{Mean} & $\mathbf{0.828}{\scriptstyle\pm0.119}$ & $\mathbf{0.822}{\scriptstyle\pm0.118}$ & $\mathbf{0.702}{\scriptstyle\pm0.095}$ & $\mathbf{0.697}{\scriptstyle\pm0.097}$ & $\mathbf{0.733}{\scriptstyle\pm0.084}$ & $\mathbf{0.726}{\scriptstyle\pm0.086}$ & $\mathbf{0.670}{\scriptstyle\pm0.073}$ & $\mathbf{0.651}{\scriptstyle\pm0.075}$ & $\mathbf{0.662}{\scriptstyle\pm0.074}$ & $\mathbf{0.644}{\scriptstyle\pm0.070}$ & $\mathbf{0.607}{\scriptstyle\pm0.052}$ & $\mathbf{0.594}{\scriptstyle\pm0.053}$ \\
\bottomrule
\end{tabular}}
\end{table}
\noindent\textit{Comparability note.} This table retains the original
regime-specific AUROCs, including prostate S3 AUROC of $0.650$.
Figure~\ref{fig:s3} and Supplementary Table~\ref{tab:supp-s3-increment}
instead use common-class, common-case evaluation for the prostate S3
comparison.
S1/S2-to-S3 age contrasts for prostate and placenta remain non-class-matched
(\suppmethodref{supp:class-comparability}).

\FloatBarrier

\FloatBarrier

\subsection*{Per-subgroup specimen counts and AUROC (Fig.~\ref{fig:results}C)}

Supplementary Table~\ref{tab:subgroup_auroc} gives the S1 and S2
cross-organ subgroup estimates and S1 training-specimen counts underlying
Fig.~\ref{fig:results}C.

\begin{table}[htbp]
  \centering\small\setlength{\tabcolsep}{6pt}
  \caption{\textbf{Per-subgroup specimen counts and mean AUROC by strategy.}
    AUROC is the macro one-vs.-rest mean across MCCV folds, averaged across all
    25 organs.
    For each demographic subgroup, S1 training specimen counts are summed
    across eligible organs within each fold and averaged over the ten MCCV folds.
    $\Delta$ = S2 $-$ S1 mean AUROC.}
  \label{tab:subgroup_auroc}
  \resizebox{\textwidth}{!}{%
\begin{tabular}{llrrrr}
    \toprule
    \textbf{Demographic} & \textbf{Subgroup} &
    \textbf{S1 training specimens (all eligible organs)} &
    \textbf{S1 AUROC} &
    \textbf{S2 AUROC} &
    \textbf{$\Delta$ (S2$-$S1)} \\
    \midrule
    \multirow{6}{*}{Age group}
      & 20--34   & 103{,}342 & 0.821 & 0.815 & $-$0.006 \\
      & 35--44   & 116{,}083 & 0.727 & 0.713 & $-$0.014 \\
      & 45--54   & 158{,}182 & 0.679 & 0.667 & $-$0.012 \\
      & 55--64   & 200{,}409 & 0.674 & 0.664 & $-$0.010 \\
      & 65--74   & 197{,}791 & 0.710 & 0.706 & $-$0.004 \\
      & $\geq$75 & 111{,}581 & 0.800 & 0.802 & $+$0.002 \\
    \midrule
    \multirow{5}{*}{Self-reported race}
      & Asian                     &  79{,}958 & 0.679 & 0.670 & $-$0.009 \\
      & Black or African American & 151{,}836 & 0.753 & 0.735 & $-$0.018 \\
      & Hispanic or Latino        & 175{,}836 & 0.619 & 0.593 & $-$0.026 \\
      & Other                     &  45{,}402 & 0.538 & 0.524 & $-$0.013 \\
      & White                     & 356{,}882 & 0.720 & 0.698 & $-$0.022 \\
    \bottomrule
  \end{tabular}}
\end{table}

\subsection*{Additional breast diagnosis-context results}
\label{supp:breast-detail}

Across the ten canonical breast diagnosis clusters, case counts ranged from
$2,681$ ($5.8\%$; benign skin/excision patterns) to $7,745$ ($16.8\%$;
benign diagnoses with calcifications). Composition used 46,145 unique cases,
including 45,750 with known age and 29,343 with known self-reported race.
All ten clusters contributed to each subgroup correlation. Relative to pooled AUROC,
cluster-stratified age-group AUROC changed by $-0.0076\pm0.0007$ under S1 and
$-0.0075\pm0.0007$ under S2; self-reported-race AUROC changed by
$-0.0075\pm0.0012$ and $-0.0080\pm0.0012$, respectively (mean$\pm$SD across
ten folds; paired $t$-test, $p<0.001$ for each;
Fig.~\ref{fig:stratification}D).

Across the ten canonical clusters, signed subgroup enrichment and
within-cluster AUROC under S1 were negatively correlated for all six age
groups, most strongly for ages 55--64 ($\rho=-0.794$) and 65--74
($\rho=-0.685$). Correlations were positive for White ($\rho=0.515$),
Asian ($\rho=0.321$), Hispanic or Latino ($\rho=0.309$), and Black or
African American ($\rho=0.152$), and negative for Other ($\rho=-0.455$;
all $n=10$ clusters). Leave-one-cluster-out ranges crossed zero for ages
20--34 and 45--54 and for Black or African American. Supplementary
Fig.~\ref{sfig:breast-enrichment-vs-auroc} shows the per-cluster values
underlying Fig.~\ref{fig:stratification}E.

\subsection*{Diagnosis-cluster quality and demographic association}

Across the 25 organs, Supplementary Table~\ref{tab:supp-cluster-quality}
reports the silhouette-selected $K$, silhouette score, maximum inter-centroid
cosine similarity, and associations between cluster membership and demographic
labels (Cram\'er's $V$). The composite criterion recommended fewer clusters
in some organs with closely related diagnosis-text clusters.

\begin{table}[htbp]
  \centering
  \caption{\textbf{Diagnosis-cluster quality and cluster--demographic
    association.}
    Specimen-level diagnosis embeddings were clustered by $k$-means per organ.
    \textbf{Clustered specimens} counts unique specimens entering organ-specific clustering. \textbf{Sel.} and
    \textbf{Rec.} are the silhouette-selected and composite-recommended $K$,
    where the composite is silhouette $\times$ (1 $-$ maximum inter-centroid
    cosine similarity, \textbf{Max sim.}). Cram\'{e}r's $V$ measures the
    association between cluster and each demographic ($\chi^2$ test at the
    selected $K$); all associations have unadjusted $p<0.05$ except those marked
    $^{\text{n.s.}}$. Sex is omitted (---) for sex-specific organs.}
  \label{tab:supp-cluster-quality}
  \scriptsize
  \setlength{\tabcolsep}{4pt}
  \resizebox{\textwidth}{!}{%
  \begin{tabular}{lrrrrrrrr}
    \toprule
     & & \multicolumn{2}{c}{\textbf{$K$}} & & & \multicolumn{3}{c}{\textbf{Cluster--demographic $V$}} \\
    \cmidrule(lr){3-4}\cmidrule(lr){7-9}
    \textbf{Organ} & \textbf{Clustered specimens} & \textbf{Sel.} & \textbf{Rec.} & \textbf{Silhouette} & \textbf{Max sim.} & \textbf{Race} & \textbf{Age} & \textbf{Sex} \\
    \midrule
Bladder      & 5,844 & 8 & 8 & 0.154 & 0.946 & 0.077 & 0.135 & 0.152 \\
Bone Marrow  & 29,563 & 10 & 10 & 0.177 & 0.966 & 0.105 & 0.121 & 0.028 \\
Bone/Joint   & 16,869 & 12 & 8 & 0.205 & 0.898 & 0.133 & 0.206 & 0.238 \\
Brain        & 4,144 & 8 & 8 & 0.244 & 0.903 & 0.119 & 0.143 & 0.144 \\
Breast       & 75,781 & 10 & 10 & 0.144 & 0.934 & 0.053 & 0.152 & --- \\
Cervix       & 30,815 & 15 & 15 & 0.231 & 0.946 & 0.043 & 0.110 & --- \\
Colon/Rectum & 144,880 & 100 & 12 & 0.469 & 0.979 & 0.149 & 0.246 & 0.154 \\
Esophagus    & 31,458 & 25 & 15 & 0.391 & 0.965 & 0.111 & 0.169 & 0.187 \\
Gallbladder & 5,562 & 6 & 6 & 0.229 & 0.953 & 0.077 & 0.083 & 0.133 \\
Gastric      & 67,104 & 10 & 15 & 0.255 & 0.970 & 0.155 & 0.138 & 0.080 \\
Kidney       & 9,606 & 8 & 8 & 0.278 & 0.885 & 0.158 & 0.146 & 0.102 \\
Liver        & 4,286 & 15 & 8 & 0.163 & 0.943 & 0.145 & 0.211 & 0.173 \\
Lung         & 4,091 & 6 & 6 & 0.187 & 0.937 & 0.124 & 0.130 & 0.124 \\
Lymph Node   & 28,473 & 10 & 8 & 0.198 & 0.939 & 0.097 & 0.120 & 0.500 \\
Mouth        & 12,953 & 10 & 10 & 0.260 & 0.869 & 0.149 & 0.248 & 0.119 \\
Ovary        & 3,487 & 6 & 6 & 0.205 & 0.905 & 0.091 & 0.309 & --- \\
Pancreas     & 1,008 & 6 & 4 & 0.195 & 0.931 & 0.084$^{\text{n.s.}}$ & 0.171 & 0.074$^{\text{n.s.}}$ \\
Placenta     & 14,795 & 12 & 10 & 0.303 & 0.943 & 0.067 & 0.043 & --- \\
Prostate     & 58,135 & 20 & 12 & 0.432 & 0.969 & 0.087 & 0.072 & --- \\
Sinus        & 11,772 & 20 & 20 & 0.276 & 0.920 & 0.147 & 0.192 & 0.100 \\
Skin         & 41,887 & 10 & 10 & 0.211 & 0.871 & 0.224 & 0.211 & 0.216 \\
Small Bowel  & 37,098 & 100 & 15 & 0.538 & 0.994 & 0.167 & 0.188 & 0.137 \\
Soft Tissue  & 13,684 & 12 & 12 & 0.168 & 0.900 & 0.107 & 0.115 & 0.142 \\
Thyroid      & 4,355 & 6 & 6 & 0.204 & 0.926 & 0.143 & 0.070 & 0.135 \\
Uterus       & 22,915 & 12 & 12 & 0.171 & 0.936 & 0.092 & 0.242 & --- \\
    \bottomrule
  \end{tabular}}
\end{table}

\clearpage
\paragraph*{Sensitivity to cohort size and cluster number}
AUROC attenuation showed no significant association with clustered-specimen
count (Spearman $\rho=+0.36$, $p=0.08$ for age group, $\rho=+0.02$ for
self-reported race, and $\rho=-0.07$ for sex) or with selected $K$
($|\rho|\leq0.24$, all $p>0.3$). Mean age-group attenuation was $-0.022$
both where selected $K$ exceeded the composite recommendation and where the
two agreed ($p=0.84$, Mann--Whitney test). Supplementary
Table~\ref{tab:supp-a4-aggregate} compares case-weighted attenuation at the
selected and recommended $K$ under S1 and S2. Supplementary
Table~\ref{tab:supp-a4-full} gives per-organ pooled and stratified AUROCs
under S1 at the selected $K$.

\Needspace{0.55\textheight}
\paragraph*{Cluster-stratified vs.\ pooled evaluation, all 69 combinations}\mbox{}\par

\begin{table}[H]
  \centering
  \caption{\textbf{Aggregate cluster-stratified evaluation.}
    Mean signed change
    $\Delta=\text{AUROC}_\text{stratified}-\text{AUROC}_\text{pooled}$ under S1
    and S2, averaged across eligible organs (age group and self-reported race,
    25 organs; sex, 19 organs). \textbf{Size-weighted} denotes the primary case-count-weighted evaluation;
    \textbf{Selected} and \textbf{Recommended} denote the $K$ used
    (Supplementary Table~\ref{tab:supp-cluster-quality}). The evaluation-only
    random-label control permutes cluster assignments for the size-weighted
    contrast. Definitions are given in the
    \suppmethodref{supp:cluster-evaluation}.}
  \label{tab:supp-a4-aggregate}
  \small
  \setlength{\tabcolsep}{6pt}
  \resizebox{\textwidth}{!}{%
    \begin{tabular}{llcccc}
    \toprule
     & & \multicolumn{3}{c}{\textbf{Size-weighted $\Delta$}} & \\
    \cmidrule(lr){3-5}
    \textbf{Strategy} & \textbf{$K$} & \textbf{Age} & \textbf{Race} & \textbf{Sex} & \textbf{Eval.\ random-label $\Delta$} \\
    \midrule
    S1 & Selected & -0.0216 & -0.0121 & -0.0095 & +0.0006 \\
    S1 & Recommended & -0.0201 & -0.0114 & -0.0086 & +0.0005 \\
    S2 & Selected & -0.0190 & -0.0118 & -0.0098 & +0.0009 \\
    S2 & Recommended & -0.0175 & -0.0113 & -0.0089 & +0.0008 \\
    \bottomrule

\end{tabular}}
\end{table}

\begin{table}[htbp]
  \centering
  \caption{\textbf{Per-organ pooled versus cluster-stratified AUROC (S1,
    selected $K$).}
    \textbf{Cl.-strat.} is the size-weighted mean of within-cluster AUROC;
    $\Delta=\text{Cl.-strat.}-\text{Pooled}$. \textbf{Drop\,\%}
    $=\Delta/(\text{Pooled}-0.5)\times100$ is secondary and unstable for
    near-chance rows (Pooled $<0.60$). Significance is from paired $t$-tests on
    $\Delta$ across 10 MCCV folds ($^{\ast}p<0.05$, $^{\ast\ast}p<0.01$,
    $^{\ast\ast\ast}p<0.001$; $^{\text{n.s.}}$ not significant; unadjusted and exploratory because MCCV folds overlap).
    Dashes (---) mark sex-specific organs excluded from the sex target.}
  \label{tab:supp-a4-full}
  \scriptsize
  \setlength{\tabcolsep}{3pt}
  \resizebox{\textwidth}{!}{%
  \begin{tabular}{lcccccccccc}
    \toprule
     & \multicolumn{3}{c}{\textbf{Age}} & \multicolumn{3}{c}{\textbf{Race}} & \multicolumn{3}{c}{\textbf{Sex}} \\
    \cmidrule(lr){2-4}\cmidrule(lr){5-7}\cmidrule(lr){8-10}
    \textbf{Organ} & Pooled & Cl.-strat. & Drop \% & Pooled & Cl.-strat. & Drop \% & Pooled & Cl.-strat. & Drop \% \\
    \midrule
Bladder      & 0.656 & 0.610 & -29.7\%*** & 0.566 & 0.559 & -10.8\%*** & 0.762 & 0.745 & -6.2\%*** \\
Bone Marrow  & 0.687 & 0.666 & -11.1\%*** & 0.645 & 0.638 & -5.0\%*** & 0.773 & 0.773 & +0.0\%$^{\text{n.s.}}$ \\
Bone/Joint   & 0.709 & 0.667 & -19.9\%*** & 0.637 & 0.612 & -18.2\%*** & 0.759 & 0.734 & -9.7\%*** \\
Brain        & 0.601 & 0.584 & -16.8\%** & 0.540 & 0.525 & -37.6\%** & 0.604 & 0.588 & -15.8\%*** \\
Breast       & 0.882 & 0.874 & -2.0\%*** & 0.750 & 0.743 & -3.0\%*** & --- & --- & --- \\
Cervix       & 0.834 & 0.827 & -2.3\%*** & 0.647 & 0.647 & -0.1\%$^{\text{n.s.}}$ & --- & --- & --- \\
Colon/Rectum & 0.784 & 0.751 & -11.7\%*** & 0.759 & 0.749 & -4.0\%*** & 0.965 & 0.959 & -1.3\%*** \\
Esophagus    & 0.692 & 0.672 & -10.6\%*** & 0.682 & 0.672 & -5.3\%*** & 0.872 & 0.866 & -1.7\%*** \\
Gallbladder & 0.790 & 0.788 & -0.5\%** & 0.669 & 0.665 & -2.3\%*** & 0.949 & 0.948 & -0.1\%$^{\text{n.s.}}$ \\
Gastric      & 0.804 & 0.796 & -2.7\%*** & 0.776 & 0.763 & -4.9\%*** & 0.970 & 0.968 & -0.5\%*** \\
Kidney       & 0.682 & 0.657 & -13.4\%*** & 0.602 & 0.576 & -25.8\%*** & 0.777 & 0.759 & -6.4\%*** \\
Liver        & 0.669 & 0.604 & -38.7\%*** & 0.605 & 0.582 & -22.1\%*** & 0.787 & 0.780 & -2.3\%** \\
Lung         & 0.732 & 0.705 & -11.8\%*** & 0.626 & 0.607 & -14.8\%*** & 0.660 & 0.655 & -2.7\%* \\
Lymph Node   & 0.789 & 0.779 & -3.7\%*** & 0.715 & 0.705 & -4.6\%*** & 0.960 & 0.927 & -7.2\%*** \\
Mouth        & 0.691 & 0.659 & -16.9\%*** & 0.612 & 0.583 & -26.1\%*** & 0.812 & 0.802 & -3.3\%*** \\
Ovary        & 0.850 & 0.788 & -17.7\%*** & 0.588 & 0.583 & -5.2\%** & --- & --- & --- \\
Pancreas     & 0.545 & 0.511 & -75.5\%* & 0.526 & 0.539 & +47.4\%* & 0.594 & 0.596 & +1.8\%$^{\text{n.s.}}$ \\
Placenta     & 0.615 & 0.625 & +8.4\%$^{\text{n.s.}}$ & 0.734 & 0.726 & -3.2\%*** & --- & --- & --- \\
Prostate     & 0.765 & 0.781 & +6.1\%$^{\text{n.s.}}$ & 0.740 & 0.740 & -0.4\%$^{\text{n.s.}}$ & --- & --- & --- \\
Sinus        & 0.759 & 0.740 & -7.3\%*** & 0.632 & 0.619 & -9.4\%*** & 0.879 & 0.867 & -3.1\%*** \\
Skin         & 0.766 & 0.744 & -8.2\%*** & 0.793 & 0.762 & -10.8\%*** & 0.912 & 0.905 & -1.7\%*** \\
Small Bowel  & 0.749 & 0.733 & -6.5\%*** & 0.726 & 0.711 & -6.6\%*** & 0.942 & 0.936 & -1.3\%*** \\
Soft Tissue  & 0.659 & 0.649 & -6.1\%*** & 0.605 & 0.588 & -16.9\%*** & 0.823 & 0.814 & -2.9\%*** \\
Thyroid      & 0.770 & 0.769 & -0.6\%$^{\text{n.s.}}$ & 0.714 & 0.701 & -6.2\%** & 0.931 & 0.929 & -0.5\%* \\
Uterus       & 0.850 & 0.814 & -10.3\%*** & 0.653 & 0.648 & -3.6\%*** & --- & --- & --- \\
    \bottomrule
  \end{tabular}}
\end{table}
\FloatBarrier

\clearpage
\subsection*{Per-organ AUROC differences between grouped-training regimes}

\begin{table}[H]
\centering
\caption{\textbf{Per-organ AUROC differences between
  S3\textsubscript{RND} and S3.} Difference $=$
  AUROC(S3\textsubscript{RND})$-$AUROC(S3), mean across 10
  matched MCCV folds; positive values indicate lower AUROC under
  diagnosis-context stratification. Significance is from paired $t$-tests on
  fold-level differences ($^{\ast}p<0.05$, $^{\ast\ast}p<0.01$,
  $^{\ast\ast\ast}p<0.001$; $^{\text{n.s.}}$ not significant); $p$-values are
  unadjusted; BH-adjusted values for this comparison appear in
  Supplementary Table~\ref{tab:supp-regime-statistics}
  (52 of 69 combinations, 75\%, have unadjusted $p<0.05$:
  22/25 age, 22/25 self-reported race, and 8/19 sex; all 52 also have
  within-target BH $q<0.05$). Fold dependence remains uncorrected, so
  these tests are exploratory. Dashes (---) mark
  sex-specific organs excluded from the sex target. Prostate age uses
  common-class, common-case evaluation; all other differences retain their
  original definitions (\suppmethodref{supp:class-comparability}).}
\label{tab:supp-s3-increment}
\scriptsize
\setlength{\tabcolsep}{4pt}
\begin{tabular}{lccc}
\toprule
\textbf{Organ} & \textbf{Age Group} & \textbf{Self-reported Race} & \textbf{Sex} \\
\midrule
Bladder & +0.0116* & +0.0066** & +0.0070** \\
Bone Marrow & +0.0099*** & +0.0070** & $-$0.0000$^{\text{n.s.}}$ \\
Bone/Joint & +0.0455*** & +0.0310*** & +0.0350*** \\
Brain & +0.0165*** & +0.0184*** & +0.0199*** \\
Breast & +0.0016*** & +0.0041*** & --- \\
Cervix & +0.0050*** & +0.0011$^{\text{n.s.}}$ & --- \\
Colon/Rectum & +0.0311*** & +0.0095*** & $-$0.0004$^{\text{n.s.}}$ \\
Esophagus & +0.0151*** & +0.0072*** & $-$0.0011$^{\text{n.s.}}$ \\
Gallbladder & $-$0.0012$^{\text{n.s.}}$ & +0.0033** & +0.0003$^{\text{n.s.}}$ \\
Gastric & +0.0076*** & +0.0068*** & $-$0.0034** \\
Kidney & +0.0296*** & +0.0340*** & +0.0010$^{\text{n.s.}}$ \\
Liver & +0.0440*** & +0.0255*** & +0.0012$^{\text{n.s.}}$ \\
Lung & +0.0079** & +0.0157*** & +0.0050** \\
Lymph Node & +0.0104*** & +0.0065*** & +0.0273*** \\
Mouth & +0.0332*** & +0.0414*** & +0.0026$^{\text{n.s.}}$ \\
Ovary & +0.0340*** & +0.0050* & --- \\
Pancreas & +0.0199** & +0.0007$^{\text{n.s.}}$ & $-$0.0008$^{\text{n.s.}}$ \\
Placenta & +0.0017$^{\text{n.s.}}$ & +0.0008$^{\text{n.s.}}$ & --- \\
Prostate & +0.0140*** & +0.0050** & --- \\
Sinus & +0.0153*** & +0.0142*** & +0.0005$^{\text{n.s.}}$ \\
Skin & +0.0155*** & +0.0298*** & $-$0.0012$^{\text{n.s.}}$ \\
Small Bowel & +0.0153*** & +0.0182*** & $-$0.0061*** \\
Soft Tissue & +0.0161*** & +0.0108*** & +0.0096*** \\
Thyroid & $-$0.0008$^{\text{n.s.}}$ & +0.0101** & $-$0.0013$^{\text{n.s.}}$ \\
Uterus & +0.0151*** & +0.0042*** & --- \\
\bottomrule
\end{tabular}
\end{table}
\FloatBarrier

\FloatBarrier
\clearpage
\subsection*{Exploratory paired comparisons across training regimes}

MCCV validation sets overlapped across folds (mean pairwise Jaccard index,
$0.111$), so fold-level differences were not independent. BH adjustment
addresses multiplicity but not fold dependence; the paired tests are
exploratory comparisons rather than confirmatory evidence.

\begingroup\scriptsize\setlength{\tabcolsep}{3pt}
\begin{longtable}{lrrrrrrrr}
\caption{\textbf{Exploratory paired training-regime comparisons.} Mean AUROC differences across ten matched MCCV folds, the number of folds with a positive difference ($n_+$), nominal paired-$t$ $p$-values, and within-target BH $q$-values. $\Delta_2=\mathrm{S2}-\mathrm{S1}$ uses the native Figure~\ref{fig:results} convention; $\Delta_3=\mathrm{AUROC}(\mathrm{S3}_{\mathrm{RND}})-\mathrm{AUROC}(\mathrm{S3})$ uses the finalized Figure~\ref{fig:s3} comparison, including common-class, common-case prostate and placenta age evaluation. Positive $\Delta_3$ indicates lower AUROC under cluster-constrained training, matching Figure~\ref{fig:s3} and Supplementary Table~\ref{tab:supp-s3-increment}; $n_+$ counts strictly positive fold differences (exact zeros are excluded).}\label{tab:supp-regime-statistics}\\
\toprule
 & \multicolumn{4}{c}{S2 relative to S1} & \multicolumn{4}{c}{S3$_{\mathrm{RND}}$ minus S3}\\
\cmidrule(lr){2-5}\cmidrule(lr){6-9}
Organ & $\Delta_2$ & $n_+$ & $p$ & $q$ & $\Delta_3$ & $n_+$ & $p$ & $q$\\\midrule
\endfirsthead
\multicolumn{9}{l}{\textit{Supplementary Table \thetable\ continued}}\\\toprule
Organ & $\Delta_2$ & $n_+$ & $p$ & $q$ & $\Delta_3$ & $n_+$ & $p$ & $q$\\\midrule
\endhead
\bottomrule\endfoot
\multicolumn{9}{l}{\textbf{Sex}}\\
Bladder & -0.0004 & 5 & 0.840 & 0.840 & +0.0070 & 9 & 0.008 & 0.019 \\
Bone Marrow & -0.0068 & 1 & 0.003 & 0.006 & -0.0000 & 6 & 0.998 & 0.998 \\
Bone/Joint & -0.0128 & 0 & 2.1e-05 & 5.7e-05 & +0.0350 & 10 & 1.1e-06 & 1.1e-05 \\
Brain & -0.0013 & 4 & 0.404 & 0.481 & +0.0199 & 10 & 1.5e-04 & 5.5e-04 \\
Colon/Rectum & -0.0024 & 0 & 2.3e-06 & 8.7e-06 & -0.0004 & 5 & 0.674 & 0.822 \\
Esophagus & -0.0206 & 0 & 7.0e-10 & 1.3e-08 & -0.0011 & 3 & 0.460 & 0.794 \\
Gallbladder & +0.0022 & 8 & 0.044 & 0.069 & +0.0003 & 6 & 0.779 & 0.822 \\
Gastric & -0.0097 & 0 & 2.6e-05 & 6.2e-05 & -0.0034 & 1 & 0.001 & 0.004 \\
Kidney & -0.0009 & 5 & 0.499 & 0.558 & +0.0010 & 6 & 0.595 & 0.822 \\
Liver & +0.0014 & 8 & 0.399 & 0.481 & +0.0012 & 6 & 0.692 & 0.822 \\
Lung & +0.0024 & 8 & 0.039 & 0.067 & +0.0050 & 9 & 0.006 & 0.018 \\
Lymph Node & +0.0004 & 5 & 0.282 & 0.412 & +0.0273 & 10 & 3.9e-08 & 7.5e-07 \\
Mouth & -0.0207 & 0 & 1.2e-05 & 3.8e-05 & +0.0026 & 6 & 0.324 & 0.615 \\
Pancreas & +0.0011 & 4 & 0.568 & 0.600 & -0.0008 & 4 & 0.736 & 0.822 \\
Sinus & +0.0010 & 5 & 0.405 & 0.481 & +0.0005 & 5 & 0.776 & 0.822 \\
Skin & -0.0127 & 0 & 2.7e-07 & 2.6e-06 & -0.0012 & 4 & 0.298 & 0.615 \\
Small Bowel & -0.0166 & 0 & 5.2e-07 & 3.3e-06 & -0.0061 & 0 & 1.3e-04 & 5.5e-04 \\
Soft Tissue & -0.0180 & 0 & 1.6e-06 & 7.6e-06 & +0.0096 & 10 & 1.2e-04 & 5.5e-04 \\
Thyroid & +0.0049 & 9 & 0.024 & 0.047 & -0.0013 & 4 & 0.643 & 0.822 \\
\addlinespace
\multicolumn{9}{l}{\textbf{Age group}}\\
Bladder & -0.0422 & 0 & 1.1e-05 & 4.7e-05 & +0.0116 & 9 & 0.027 & 0.030 \\
Bone Marrow & -0.0215 & 0 & 5.3e-06 & 3.3e-05 & +0.0099 & 10 & 2.1e-06 & 4.1e-06 \\
Bone/Joint & -0.0092 & 0 & 9.1e-06 & 4.6e-05 & +0.0455 & 10 & 1.2e-08 & 5.1e-08 \\
Brain & -0.0015 & 3 & 0.164 & 0.241 & +0.0165 & 10 & 1.5e-05 & 2.4e-05 \\
Breast & -0.0036 & 0 & 1.4e-06 & 1.6e-05 & +0.0016 & 10 & 2.5e-04 & 3.3e-04 \\
Cervix & -0.0190 & 0 & 6.7e-07 & 1.6e-05 & +0.0050 & 10 & 2.2e-05 & 3.3e-05 \\
Colon/Rectum & -0.0086 & 0 & 6.5e-05 & 2.0e-04 & +0.0311 & 10 & 4.8e-13 & 1.2e-11 \\
Esophagus & +0.0028 & 10 & 2.9e-05 & 1.0e-04 & +0.0151 & 10 & 1.2e-08 & 5.1e-08 \\
Gallbladder & +0.0001 & 5 & 0.850 & 0.859 & -0.0012 & 3 & 0.144 & 0.157 \\
Gastric & -0.0000 & 6 & 0.859 & 0.859 & +0.0076 & 10 & 2.9e-08 & 9.0e-08 \\
Kidney & +0.0010 & 5 & 0.384 & 0.505 & +0.0296 & 10 & 1.2e-06 & 2.4e-06 \\
Liver & -0.0019 & 3 & 0.043 & 0.082 & +0.0440 & 10 & 2.5e-07 & 5.8e-07 \\
Lung & -0.0160 & 1 & 8.2e-04 & 0.002 & +0.0079 & 8 & 0.004 & 0.005 \\
Lymph Node & -0.0031 & 2 & 0.149 & 0.232 & +0.0104 & 10 & 9.8e-08 & 2.7e-07 \\
Mouth & -0.0004 & 4 & 0.378 & 0.505 & +0.0332 & 10 & 2.7e-08 & 9.0e-08 \\
Ovary & -0.0056 & 1 & 0.004 & 0.008 & +0.0340 & 10 & 2.8e-06 & 5.0e-06 \\
Pancreas & +0.0039 & 6 & 0.693 & 0.788 & +0.0199 & 8 & 0.004 & 0.004 \\
Placenta & -0.0385 & 3 & 0.063 & 0.112 & +0.0017 & 6 & 0.229 & 0.238 \\
Prostate & -0.0145 & 5 & 0.501 & 0.626 & +0.0140 & 10 & 1.5e-05 & 2.4e-05 \\
Sinus & -0.0097 & 0 & 0.002 & 0.004 & +0.0153 & 10 & 4.0e-05 & 5.6e-05 \\
Skin & +0.0000 & 4 & 0.834 & 0.859 & +0.0155 & 10 & 2.9e-10 & 3.6e-09 \\
Small Bowel & +0.0003 & 5 & 0.660 & 0.786 & +0.0153 & 10 & 1.7e-07 & 4.2e-07 \\
Soft Tissue & +0.0048 & 10 & 0.001 & 0.003 & +0.0161 & 10 & 6.0e-09 & 3.7e-08 \\
Thyroid & +0.0035 & 8 & 0.113 & 0.188 & -0.0008 & 3 & 0.278 & 0.278 \\
Uterus & -0.0084 & 0 & 1.9e-06 & 1.6e-05 & +0.0151 & 10 & 5.7e-09 & 3.7e-08 \\
\addlinespace
\multicolumn{9}{l}{\textbf{Self-reported race}}\\
Bladder & -0.0181 & 0 & 2.7e-04 & 3.9e-04 & +0.0066 & 9 & 0.001 & 0.002 \\
Bone Marrow & -0.0171 & 0 & 9.4e-06 & 2.6e-05 & +0.0070 & 9 & 0.007 & 0.009 \\
Bone/Joint & -0.0309 & 0 & 1.7e-06 & 1.1e-05 & +0.0310 & 10 & 1.4e-07 & 8.6e-07 \\
Brain & -0.0040 & 3 & 0.129 & 0.134 & +0.0184 & 10 & 1.9e-06 & 5.9e-06 \\
Breast & -0.0236 & 0 & 1.1e-06 & 9.2e-06 & +0.0041 & 10 & 2.5e-06 & 6.9e-06 \\
Cervix & -0.0149 & 0 & 2.4e-05 & 5.6e-05 & +0.0011 & 7 & 0.121 & 0.131 \\
Colon/Rectum & -0.0224 & 0 & 9.3e-06 & 2.6e-05 & +0.0095 & 10 & 1.1e-06 & 3.8e-06 \\
Esophagus & -0.0228 & 0 & 6.8e-06 & 2.4e-05 & +0.0072 & 10 & 1.1e-05 & 2.8e-05 \\
Gallbladder & -0.0090 & 1 & 0.004 & 0.005 & +0.0033 & 9 & 0.005 & 0.006 \\
Gastric & -0.0114 & 0 & 2.6e-05 & 5.6e-05 & +0.0068 & 10 & 1.4e-05 & 3.2e-05 \\
Kidney & -0.0128 & 1 & 4.5e-04 & 6.2e-04 & +0.0340 & 10 & 3.9e-08 & 3.2e-07 \\
Liver & -0.0054 & 1 & 0.004 & 0.005 & +0.0255 & 10 & 7.7e-07 & 3.2e-06 \\
Lung & -0.0131 & 1 & 6.7e-04 & 8.8e-04 & +0.0157 & 10 & 3.8e-05 & 7.3e-05 \\
Lymph Node & -0.0292 & 0 & 3.7e-06 & 1.9e-05 & +0.0065 & 10 & 8.5e-05 & 1.4e-04 \\
Mouth & -0.0208 & 0 & 3.3e-05 & 6.4e-05 & +0.0414 & 10 & 2.4e-07 & 1.2e-06 \\
Ovary & -0.0108 & 2 & 0.018 & 0.019 & +0.0050 & 8 & 0.034 & 0.038 \\
Pancreas & -0.0083 & 4 & 0.300 & 0.300 & +0.0007 & 5 & 0.879 & 0.879 \\
Placenta & -0.0293 & 0 & 1.1e-09 & 2.8e-08 & +0.0008 & 5 & 0.431 & 0.449 \\
Prostate & -0.0339 & 0 & 7.2e-05 & 1.2e-04 & +0.0050 & 8 & 0.006 & 0.007 \\
Sinus & -0.0159 & 1 & 2.4e-04 & 3.7e-04 & +0.0142 & 10 & 2.7e-05 & 5.7e-05 \\
Skin & -0.0260 & 0 & 7.9e-09 & 9.9e-08 & +0.0298 & 10 & 4.0e-10 & 9.3e-09 \\
Small Bowel & -0.0176 & 0 & 2.7e-05 & 5.6e-05 & +0.0182 & 10 & 7.5e-10 & 9.3e-09 \\
Soft Tissue & -0.0221 & 0 & 4.1e-05 & 7.3e-05 & +0.0108 & 10 & 4.9e-05 & 8.7e-05 \\
Thyroid & -0.0141 & 1 & 0.004 & 0.005 & +0.0101 & 9 & 0.001 & 0.002 \\
Uterus & -0.0096 & 0 & 4.7e-06 & 2.0e-05 & +0.0042 & 10 & 3.5e-04 & 5.5e-04 \\
\addlinespace
\end{longtable}
\endgroup

\end{document}